\documentclass{article}

\PassOptionsToPackage{numbers, compress}{natbib}

\usepackage[preprint]{neurips_2020}

\usepackage[utf8]{inputenc} \usepackage[T1]{fontenc}    \usepackage{hyperref}       \usepackage{url}            \usepackage{booktabs}       \usepackage{amsfonts}       \usepackage{nicefrac}       \usepackage{microtype}      

\usepackage{color}
\usepackage{epsfig}
\usepackage{amsthm}
\usepackage{amsmath}
\usepackage{float}
\usepackage{graphicx}
\usepackage{wrapfig}
\usepackage{makecell}
\usepackage{caption}
\usepackage{subcaption}
\usepackage{bm}
\usepackage{multirow}
\usepackage{mathtools}

\title{Are Powerful Graph Neural Nets Necessary? \\A Dissection on Graph Classification}

\author{Ting Chen\thanks{Now at Google.} \\
    University of California, Los Angeles\\
\texttt{tingchen@cs.ucla.edu} \\
\And
    Song Bian \\
    Zhejiang University \\
    \texttt{songbian@zju.edu.cn} \\
\And
    Yizhou Sun \\
    University of California, Los Angeles\\
\texttt{yzsun@cs.ucla.edu} \\
}

\newtheorem{claim}{Claim}
\newtheorem{proposition}{Proposition}\theoremstyle{definition}
\newtheorem{definition}{Definition}

\newcommand{\nop}[1]{}
\newcommand{\wappendix}{appendix}

\begin{document}
\maketitle

\begin{abstract}
Graph Neural Nets (GNNs) have received increasing attentions, partially due to their superior performance in many node and graph classification tasks. However, there is a lack of understanding on what they are learning and how sophisticated the learned graph functions are. 
In this work, we propose a dissection of GNNs on graph classification into two parts: 1) the graph filtering, where graph-based neighbor aggregations are performed, and 2) the set function, where a set of hidden node features are composed for prediction. To study the importance of both parts, we propose to linearize them separately. We first linearize the graph filtering function, resulting Graph Feature Network (GFN), which is a simple lightweight neural net defined on a \textit{set} of graph augmented features. Further linearization of GFN's set function results in Graph Linear Network (GLN), which is a linear function.
Empirically we perform evaluations on common graph classification benchmarks. To our surprise, we find that, despite the simplification, GFN could match or exceed the best accuracies produced by recently proposed GNNs (with a fraction of computation cost), while GLN underperforms significantly. Our results demonstrate the importance of non-linear set function, and suggest that linear graph filtering with non-linear set function is an efficient and powerful scheme for modeling existing graph classification benchmarks.\footnote{Code at: \href{https://github.com/chentingpc/gfn}{https://github.com/chentingpc/gfn}.}
\end{abstract}

 \vspace{-1em}
\section{Introduction}

Recent years have seen increasing attention to Graph Neural Nets (GNNs)~\cite{scarselli2009graph,li2015gated,defferrard2016convolutional,kipf2016semi}, which have achieved superior performance in many graph tasks, such as node classification~\cite{kipf2016semi,wu2019simplifying} and graph classification~\cite{simonovsky2017dynamic,xinyi2018capsule}. Different from traditional neural networks that are defined on regular structures such as sequences or images, graphs provide a more general abstraction for structured data, which subsume regular structures as special cases. The power of GNNs is that they can directly define learnable compositional function on (arbitrary) graphs, thus extending classic networks (e.g. CNNs, RNNs) to more irregular and general domains.

Despite their success, it is unclear what GNNs have learned, and how sophisticated the learned graph functions are. It is shown in~\cite{zeiler2014visualizing} that traditional CNNs used in image recognition have learned complex hierarchical and compositional features, and that deep non-linear computation can be beneficial~\cite{he2016identity}. Is this also the case when applying GNNs to common graph problems? Recently, \cite{wu2019simplifying} showed that, for common node classification benchmarks, non-linearity can be removed in GNNs without suffering much loss of performance. The resulting linear GNNs collapse into a logistic regression on graph propagated features. This raises doubts on the necessity of complex GNNs, which require much more expensive computation, for node classification benchmarks. Here we take a step further dissecting GNNs, and examine the necessity of complex GNN parts on more challenging graph classification benchmarks~\cite{yanardag2015deep,zhang2018end,xinyi2018capsule}.

To better understand GNNs on graph classification, we dissect it into two parts/stages: 1) the graph filtering part, where graph-based neighbor aggregations are performed, and 2) the set function part, where a set of hidden node features are composed for prediction. We aim to test the importance of both parts separately, and seek answers to the following questions. \textit{Do we need a sophisticated graph filtering function for a particular task or dataset? And if we have a powerful set function, is it enough to use a simple graph filtering function?}

To answer these questions, we propose to linearize both parts separately. We first linearize graph filtering part, resulting Graph Feature Network (GFN): a simple lightweight neural net defined on a set of graph augmented features. Unlike GNNs, which learn a multi-step neighbor aggregation function on graphs~\cite{dai2016discriminative,gilmer2017neural}, the GFN only utilizes graphs in constructing its input features. It first augments nodes with graph structural and propagated features, and then learns a neural net directly on the \textit{set} of nodes (i.e. a bag of graph pre-processed feature vectors), which make it more efficient. We then further linearize set function in GFN, and arrive at Graph Linear Network (GLN), which is a linear function of augmented graph features.

Empirically, we perform evaluations on common graph classification benchmarks~\cite{yanardag2015deep,zhang2018end,xinyi2018capsule}, and find that GFN can match or exceed the best accuracies produced by recently proposed GNNs, at a fraction of the computation cost. GLN performs much poorly than both GFN and recent GNNs. This result casts doubts on the necessity of non-linear graph filtering, and suggests that the existing GNNs may not have learned more sophisticated graph functions than linear neighbor aggregation on these benchmarks. Furthermore, we find non-linear set function plays an important role, as its linearization can hurt performance significantly.

 \vspace{-1em}
\section{Preliminaries}

\textbf{Graph classification problem.} We use $G = (V, E)\in\mathcal{G}$ to denote a graph, where $V$ is a set of vertices/nodes, and $E$ is a set of edges. We further denote an attributed graph as $G_X = (G, X) \in \mathcal{G}_X$, where $X\in\mathbb{R}^{n\times d}$ are node attributes with $n=|V|$. It is assumed that each attributed graph is associated with some label $y\in \mathcal{Y}$, where $\mathcal{Y}$ is a set of pre-defined categories. The goal in graph classification problem is to learn a mapping function $f:\mathcal{G}_X \rightarrow \mathcal{Y}$, such that we can predict the target class for unseen graphs accurately. Many real world problems can be formulated as graph classification problems, such as social and biological graph classification~\cite{yanardag2015deep,kipf2016semi}.

\textbf{Graph neural networks.}
Graph Neural Networks (GNNs) define functions on the space of attributed graph $\mathcal{G}_X$. Typically, the graph function, $\text{GNN}(G, X)$, learns a multiple-step transformation of the original attributes/signals for final node level or graph level prediction. In each of the step $t$, a new node presentation, $h^{(t)}_v$ is learned. Initially, $h^{(1)}_v$ is initialized with the node attribute vector, and during each subsequent step, a \textit{neighbor aggregation function} is applied to generate the new node representation. More specifically, common neighbor aggregation functions for the $v$-th node take the following form:
\begin{equation}
\label{eq:aggregation}
h^{(t)}_v = f\bigg(h^{(t-1)}_v, \bigg\{h_u^{(t-1)}| u \in \mathcal{N}(v)\bigg\} \bigg),
\end{equation}
where $\mathcal{N}(v)$ is a set of neighboring nodes of node $v$.
To instantiate this neighbor aggregation function, \cite{kipf2016semi} proposes the Graph Convolutional Network (GCN) aggregation scheme as follows.
\begin{equation}
\label{eq:gcn}
h_v^{(t+1)} = \sigma \bigg(\sum_{u \in \mathcal{N}(v)} \tilde{A}_{uv} (W^{(t)})^Th_u^{(t)} \bigg),
\end{equation}
where $W^{(t)}\in \mathbb{R}^{d\times d'}$ is the learnable transformation weight, $\tilde{A} = \tilde{D}^{-1/2}(A+\epsilon I)\tilde{D}^{-1/2}$ is the normalized adjacency matrix with $\epsilon$ as a constant ($\epsilon=1$ in~\cite{kipf2016semi}) and $\tilde{D}_{ii} = \sum_j A_{ij}+\epsilon$. $\sigma(\cdot)$ is a non-linear activation function, such as ReLU. This transformation can also be written as $H^{(t+1)} = \sigma (\tilde{A} H^{(t)} W^{(t)})$, where $H^{(t)}\in \mathbb{R}^{n\times d}$ are the hidden states of all nodes at $t$-th step.

More sophisticated neighbor aggregation schemes are also proposed, such as GraphSAGE~\cite{hamilton2017inductive} which allows pooling and recurrent aggregation over neighboring nodes. Most recently, in Graph Isomorphism Network (GIN)~\cite{xu2018how}, a more powerful aggregation function is proposed as follows.
\begin{equation}
\label{eq:gin}
h^{(t)}_v = \text{MLP}^{(t)}\bigg(\bigg(1+\epsilon^{(t)}\bigg) h^{(t-1)}_v + \sum_{u\in \mathcal{N}(v)} h^{(t-1)}_u\bigg),
\end{equation}
where MLP abbreviates for multi-layer perceptrons and $\epsilon^{(t)}$ can either be zero or a learnable parameter.

Finally, in order to generate graph level representation $h_{G}$, a \textit{readout function} is used, which generally takes the following form:
\begin{equation}
\label{eq:readout}
h_G = g\bigg(\bigg\{h_v^{(T)}|v\in G\bigg\}\bigg).
\end{equation}
This can be instantiated by a global sum pooling, i.e. $h_G = \sum_{v=1}^n h_v^{(T)}$ followed by fully connected layers to generate the categorical or numerical output. \vspace{-1em}
\section{Approach}

\subsection{Graph feature network}

Motivated by the question that, with a powerful graph readout function, whether we can simplify the sophisticated multi-step neighbor aggregation functions (such as Eq. \ref{eq:gcn} and \ref{eq:gin}). Therefore we propose Graph Feature Network (GFN): a neural set function defined on a set of graph augmented features.

\textbf{Graph augmented features.} In GFN, we replace the sophisticated neighbor aggregation functions (such as Eq. \ref{eq:gcn} and \ref{eq:gin}) with graph augmented features based on $G_X$. Here we consider two categories as follows: 1) graph structural/topological features, which are related to the intrinsic graph structure, such as node degrees, or node centrality scores\footnote{We only use node degree in this work as it is very efficient to calculate during both training and inference.},  but do not rely on node attributes; 2) graph propagated features, which leverage the graph as a medium to propagate node attributes.
The graph augmented features $X^G$ can be seen as the output of a feature extraction function defined on the attributed graph, i.e. $X^G = \gamma(G, X)$, and Eq. \ref{eq:gfn_1} below gives a specific form, which combine node degree features and multi-scale graph propagated features as follows:
\begin{equation}
\label{eq:gfn_1}
X^G = \gamma(G, X) = \bigg[\bm d, X, \tilde{A}^1X, \tilde{A}^2X, \cdots, \tilde{A}^KX\bigg],
\end{equation} 
where $\bm d\in \mathbb{R}^{n\times 1}$ is the degree vector for all nodes, and $\tilde{A}$ is again the normalized adjacency matrix ($\tilde{A} = \tilde{D}^{-1/2}(A+\epsilon I)\tilde{D}^{-1/2}$), but other designs of propagation operator are possible~\cite{klicpera2018combining}. Features separated by comma are concatenated to form $X^G$.

\textbf{Neural set function.} To build a powerful graph readout function based on graph augmented features $X^G$, we use a neural set function. The neural set function discards the graph structures and learns purely based on the set of augmented node features. Motivated by the general form of a permutation-invariant set function shown in~\cite{zaheer2017deep}, we define our neural set function for GFN as follows:
\begin{equation}
\label{eq:gfn_2}
\text{GFN}(G, X)=\rho\bigg(\sum_{v\in\mathcal{V}}\phi\bigg(X^G_v\bigg)\bigg).
\end{equation}
Both $\phi(\cdot)$ and $\rho(\cdot)$ are parameterized by neural networks. 
Concretely, we parameterize the function $\phi(\cdot)$  as a multi-layer perceptron (MLP), i.e. $\phi(x)=\sigma(\sigma(\cdots\sigma(x^TW^{(1)})\cdots) W^{(T)})$. Note that a single layer of $\phi(\cdot)$ resembles a graph convolution layer $H^{(t+1)} = \sigma(\tilde{A}H^{(t)}W^{(t)})$ with the normalized adjacency matrix $\tilde{A}$ replaced by identity matrix $I$ (a.k.a. $1\times 1$ convolution). 
As for the function $\rho(\cdot)$, we parameterize it with another MLP (i.e. fully connected layers in this case).

\textbf{Computation efficiency.} GFN provides a way to approximate GNN with less computation overheads, especially during the training process. Since the graph augmented features can be pre-computed before training starts, the graph structures are not involved in the iterative training process. This brings the following advantages. First, since there is no neighbor aggregation step in GFN, it reduces computational complexity. To see this, one can compare a single layer feature transformation function in GFN, i.e. $\sigma(HW)$, against the neighbor aggregation function in GCN, i.e. $\sigma(\tilde{A}HW)$. Secondly, since graph augmented features of different scales are readily available from the input layer, GFN can leverage them much earlier, thus may require fewer transformation layers. Lastly, it also eases the implementation related overhead, since the neighbor aggregation operation in graphs are typically implemented by sparse matrix operations.

\textbf{Graph Linear Network.} When we use a linear set function instead of the generic one used in Eq. \ref{eq:gfn_2}, we arrive at graph linear network, which can be expressed as follows.
\begin{equation}
\label{eq:gln}
\text{GLN}(G, X)=\sigma\bigg(W\sum_{v\in\mathcal{V}}\bigg(X^G_v\bigg)\bigg).
\end{equation}
Where $W$ is a weight matrix, and $\sigma(\cdot)$ is softmax function produce class probability.

\subsection{From GNN to GFN and GLN: a dissection of GNNs}
To better understand GNNs on graph classification, we propose a formal dissection/decomposition of GNNs into two parts/stages: the graph filtering part and the set function part. As we shall see shortly, the simplification of the graph filtering part allows us to derive GFN from GNN, and also be able to assess the importance of the two GNN parts separately. 

To make concepts more clear, we first give formal definitions of the two GNN parts in the dissection.
\begin{definition}
\label{th:def1}
(Graph filtering) 
A graph filtering function, $Y = \mathcal{F}_G(X)$, performs a transformation of input signals based on the graph $G$, which takes a set of signals $X\in \mathbb{R}^{n\times d}$ and outputs another set of filtered signals $Y\in \mathbb{R}^{m \times d'}$.
\end{definition}
Graph filtering in most existing GNNs consists of multi-step neighbor aggregation operations, i.e. multiple steps of Eq. \ref{eq:aggregation}. For example, in GCN~\cite{kipf2016semi}, the multi-step neighbor aggregation can be expressed as $H^{(T)} = \sigma(A\sigma(...\sigma(AXW^{(1)})...)W^{(T)})$.

\begin{definition}
\label{th:def2}
(Set function) A set function, $y=\mathcal{T}(Y)$, takes a set of vectors $Y\in \mathbb{R}^{m \times d'}$ where their order does not matter, and outputs a task specific prediction $y\in \mathcal{Y}$.
\end{definition}
The graph readout function in Eq. \ref{eq:readout} is a set function, which enables the graph level prediction that is permutation invariant w.r.t. nodes in the graph. Although a typical readout function is simply a global pooling~\cite{xu2018how}, the set function can be as complicated as Eq. \ref{eq:gfn_2}.

\begin{claim}
\label{th:claim}
A GNN that is a mapping of $\mathcal{G}_X\rightarrow \mathcal{Y}$ can be decomposed into a graph filtering function followed by a set function, i.e. $\text{GNN}(G, X) = \mathcal{T}\circ \mathcal{F}_G(X)$.
\end{claim}

This claim is obvious for the neighbor aggregation framework defined by Eq. \ref{eq:aggregation} and \ref{eq:readout}, where most existing GNN variants such as GCN, GraphSAGE and GIN follow. This claim is also general, even for unforeseen GNN variants that do not explicitly follow this framework~\footnote{For some unforeseen cases, we can absorb the set function $\mathcal{T}$ into $\mathcal{F}_G$. That is, let the output $Y=\mathcal{F}_G(\cdot)$ be final logits for pre-defined classes and set $\mathcal{T}(\cdot)$ to softmax function with zero temperature, i.e. $\exp(x/\tau)/Z$ with $\tau\rightarrow 0$.}.

We aim to assess the importance of two GNN parts separately. However, it is worth pointing out that the above decomposition is not unique in general, and the functionality of the two parts can overlap: if the graph filtering part has fully transformed graph features, then a simple set function may be used for prediction. This makes it challenging to answer the question: do we need a sophisticated graph filtering part for a particular task or dataset, especially when a powerful set function is used? To better disentangle these two parts and study their importance more independently, similar to~\cite{wu2019simplifying}, we propose to simplify the graph filtering part by linearizing it.

\begin{definition}
\label{th:def3}
(Linear graph filtering) We say a graph filtering function $\mathcal{F}_G(X)$ is linear w.r.t. $X$ iff it can be expressed as $\mathcal{F}_G(X) = \Gamma(G, X)\bm \theta$, where $\Gamma(G, X)$ is a linear map of $X$, and $\bm\theta$ is the only learnable parameter.
\end{definition}
Intuitively, one can construct a linear graph filtering by removing the non-linear operations from graph filtering part in existing GNNs, such as non-linear activation function $\sigma(\cdot)$ in Eq. \ref{eq:gcn} or \ref{eq:gin}. By doing so, the graph filtering becomes linear w.r.t. X, thus multi-layer weights collapse into a single linear transformation, described by $\bm \theta$. More concretely, let us consider a linearized GCN~\cite{kipf2016semi}, its $K$-th layer can be written as $H^{(K)}=\tilde{A}^KX(\Pi_{k={1}}^K W^{(k)})$, and we can rewrite the weights with $\bm\theta = \Pi_{k={1}}^K W^{(k)}$.

The linearization of graph filtering part enables us to disentangle graph filtering and the set function more thoroughly: the graph filtering part mainly constructs graph augmented features (by setting $\gamma(G, X) = \Gamma(G, X)$), and the set function learns to compose them for the graph-level prediction. This leads to the proposed GFN. In other words, GNNs with a linear graph filtering part can be expressed as GFN with appropriate graph augmented features. This is shown more formally in the following proposition \ref{th:prop1}.
\begin{proposition}
\label{th:prop1}
Let $\text{GNN}^{lin}(G, X)$ be a mapping of $\mathcal{G}_X \rightarrow \mathcal{Y}$ that has a linear graph filtering part, i.e. $\mathcal{F}_G(X)=\Gamma(G, X)\bm \theta$, then we have $\text{GNN}^{lin}(G, X) = \text{GFN}(G, X)$, where $\gamma(G, X) = \Gamma(G, X)$.
\end{proposition}
The proof can be found in the \wappendix. Noted that a GNN with a linear graph filtering can be seen as a GFN, but the reverse may not be true. General GFN can also have non-linear graph filtering, e.g. when the feature extraction function $\gamma(G, X)$ is not a linear map of $X$ (Eq. \ref{eq:gfn_1} is a linear map of $X$). 

\textbf{Why GFN?} GFN can help us understand the functions that GNNs learned on current benchmarks. First, by comparing GNN with linear graph filtering (i.e. GFN) against standard GNN with non-linear graph filtering, we can assess the importance of non-linear graph filtering part. Secondly, by comparing GFN with linear set function (i.e. GLN) against GFN with non-linear set function, we can assess the importance of non-linear set function. 
 \vspace{-1em}\section{Experiments}
\subsection{Datasets and settings}

\textbf{Datasets.} The main datasets we consider are commonly used graph classification benchmarks~\cite{yanardag2015deep,xinyi2018capsule,xu2018how}. The graphs in the collection can be categorized into two categories biological graphs and social graphs. It is worth noting that the social graphs have no node attributes, while the biological graphs come with categorical node attributes. The detailed statistics for biological and social graph datasets can be found in Appendix. 

\textbf{Baselines.} We compare with two families of baselines.  
The first family of baselines are kernel-based, namely the Weisfeiler-Lehman subtree kernel (WL)~\cite{shervashidze2011weisfeiler}, 
Deep Graph Kernel (DGK)~\cite{yanardag2015deep} and AWE~\cite{ivanov2018anonymous} that incorporate kernel-based methods with learning-based approach to learn embeddings. Two recent work alone this line are RetGK~\cite{zhang2018retgk} and GNTK~\cite{Du2019GraphNT}.

The second family of baselines are GNN-based models, which include recently proposed PATCHY-SAN (PSCN)~\cite{niepert2016learning}, Deep Graph CNN (DGCNN)~\cite{zhang2018end}, CapsGNN~\cite{xinyi2018capsule} and GIN~\cite{xu2018how}. GNN based methods are usually more scalable compared to graph kernel based ones, as the complexity is typically quadratic in the number of graphs and nodes for kernels while linear for GNNs.

For the above baselines, we use their accuracies reported in the original papers, following the same evaluation setting as in~\cite{xu2018how}. Architecture and hyper-parameters can make a difference, so to enable a better controlled comparison between GFN and GNN, we also implement Graph Convolutional Networks (GCN) from~\cite{kipf2016semi}. More specifically, our GCN model contains a dense feature transformation layer, i.e. $H^{(2)}=\sigma(XW^{(1)})$, followed by three GCN layers, i.e. $H^{(t+1)}=\sigma(\tilde{A}H^{(t)}W^{(t)})$. We also vary the number of GCN layers in our ablation study. To enable graph level prediction, we add a global sum pooling, followed by two fully-connected layers that produce categorical probability over pre-defined categories.

\begin{table*}[!t]
\footnotesize
\centering
\caption{\label{tab:methods_bio}Test accuracies (\%) for biological graphs. The best results per dataset and in average are highlighted. - means the results are not available for a particular dataset.}
\begin{tabular}{lcccccc}
\Xhline{1.6\arrayrulewidth}\toprule
Algorithm &          MUTAG &           NCI1 &       PROTEINS &             D\&D &        ENZYMES & \textbf{Average}\\
\midrule\midrule
WL     & 82.05$\pm$0.36  &  82.19$\pm$0.18 &  74.68$\pm$0.49 &   {79.78$\pm$0.36}   &    52.22$\pm$1.26 & 74.18\\
AWE     & 87.87$\pm$9.76  &  - &  - &   71.51$\pm$4.02   &    35.77$\pm$5.93 & - \\
DGK     & 87.44$\pm$2.72  &  80.31$\pm$0.46 &  75.68$\pm$0.54 &   73.50$\pm$1.01   &    53.43$\pm$0.91 & 74.07\\
RetGK$_{I}$ & {90.30$\pm$1.10} & \textbf{84.50$\pm$0.20} & 75.80$\pm$0.60 & \textbf{81.60$\pm$0.30} & 60.40$\pm$0.80 & {78.52} \\
RetGK$_{II}$ & 90.10$\pm$1.00 & 83.50$\pm$0.20 & 75.20$\pm$0.30 & 81.00$\pm$0.50 & 59.10$\pm$1.10 & 77.78 \\
GNTK & 90.00$\pm$8.50 & 84.20$\pm$1.50 & 75.60$\pm$4.20 & - & - & - \\
\midrule
PSCN     & 88.95$\pm$4.37  &  76.34$\pm$1.68 &  75.00$\pm$2.51 &   76.27$\pm$2.64   &   - & - \\
DGCNN    & 85.83$\pm$1.66  &  74.44$\pm$0.47 &  75.54$\pm$0.94 &   79.37$\pm$0.94   &   51.00$\pm$7.29 & 73.24\\
CapsGNN     & 86.67$\pm$6.88  &  78.35$\pm$1.55 &  76.28$\pm$3.63 &   75.38$\pm$4.17   & 54.67$\pm$5.67 & 74.27 \\
GIN  &  89.40$\pm$5.60 &  82.70$\pm$1.70 &  76.20$\pm$2.80 &  - &   - & - \\
\midrule
GCN           &  87.20$\pm$5.11 &  {83.65$\pm$1.69} &  75.65$\pm$3.24 &  79.12$\pm$3.07 &  66.50$\pm$6.91 & {78.42}\\
GLN           &  82.85$\pm$12.15 &  68.61$\pm$2.31 &  75.65$\pm$4.43 &  76.75$\pm$5.00 &  43.83$\pm$5.16 & {69.54}\\
GFN           & \textbf{90.84$\pm$7.22} &  82.77$\pm$1.49 &  76.46$\pm$4.06 &  78.78$\pm$3.49 & \textbf{70.17$\pm$5.58} & \textbf{79.80}\\
GFN-light &  89.89$\pm$7.14 &  81.43$\pm$1.65 &  \textbf{77.44$\pm$3.77} &  78.62$\pm$5.43 &  69.50$\pm$7.37 & {79.38}\\
\bottomrule\Xhline{1.6\arrayrulewidth}
\end{tabular}
\vspace{-1em}
\end{table*}

\begin{table*}[!t]
\footnotesize
\centering
\caption{\label{tab:methods_social}Test accuracies (\%) for social graphs. The best results per dataset and in average are highlighted. - means the results are not available for a particular dataset.}
\begin{tabular}{lcccccc}
\Xhline{1.6\arrayrulewidth}\toprule
Algorithm &         COLLAB &   IMDB-B &     IMDB-M & RE-M5K & RE-M12K & \textbf{Average}\\
\midrule\midrule
WL     & 79.02$\pm$1.77  &  73.40$\pm$4.63 &  49.33$\pm$4.75 &   49.44$\pm$2.36   &    38.18$\pm$1.30 & 57.87\\
AWE     & 73.93$\pm$1.94  &  74.45$\pm$5.83 &  51.54$\pm$3.61 &   50.46$\pm$1.91   &    39.20$\pm$2.09 & 57.92\\
DGK     & 73.09$\pm$0.25  &  66.96$\pm$0.56 &  44.55$\pm$0.52 &   41.27$\pm$0.18   &    32.22$\pm$0.10 & 51.62\\
RetGK$_{I}$ & 81.00$\pm$0.30 & 71.90$\pm$1.00 & 47.70$\pm$0.30 & {56.10$\pm$0.50} & {48.70$\pm$0.20} & {61.08} \\
RetGK$_{II}$ & 80.60$\pm$0.30 & 72.30$\pm$0.60 & 48.70$\pm$0.60 & 55.30$\pm$0.30 & 47.10$\pm$0.30 & 60.80 \\
GNTK &\textbf{83.60$\pm$1.00} & \textbf{76.90$\pm$3.60 }&\textbf{52.80$\pm$4.60} & - & - & - \\
\midrule
PSCN     & 72.60$\pm$2.15  &  71.00$\pm$2.29 &  45.23$\pm$2.84 &   49.10$\pm$0.70   &   41.32$\pm$0.42 & 55.85\\
DGCNN    & 73.76$\pm$0.49  &  70.03$\pm$0.86 &  47.83$\pm$0.85 &   48.70$\pm$4.54   &   -  & - \\
CapsGNN     & 79.62$\pm$0.91  &  73.10$\pm$4.83 & 50.27$\pm$2.65 &   52.88$\pm$1.48   & 46.62$\pm$1.90 & 60.50 \\
GIN   &  80.20$\pm$1.90 &  {75.10$\pm$5.10} &  {52.30$\pm$2.80} &  57.50$\pm$1.50 &   - & - \\
\midrule
GCN           &  {81.72$\pm$1.64} &  73.30$\pm$5.29 &  51.20$\pm$5.13 &   56.81$\pm$2.37 &    49.31$\pm$1.44 &   62.47 \\
GLN          &  75.72$\pm$2.51 &  73.10$\pm$3.18 &  50.40$\pm$5.61 &  52.97$\pm$2.58 &  39.84$\pm$0.95 & 58.41\\
GFN           &  81.50$\pm$2.42 &  73.00$\pm$4.35 &  51.80$\pm$5.16 &   \textbf{57.59$\pm$2.40} &    49.43$\pm$1.36 &   \textbf{62.66} \\
GFN-light &  81.34$\pm$1.73 &  73.00$\pm$4.29 &  51.20$\pm$5.71 &   57.11$\pm$1.46 &   \textbf{49.75$\pm$1.19} &   62.48 \\
\bottomrule\Xhline{1.6\arrayrulewidth}
\end{tabular}
\vspace{-1em}
\end{table*}

\textbf{Model configurations.} For the proposed GFN, we \textit{mirror} our GCN model configuration to allow direct comparison. Therefore, we use the same architecture, parameterization and training setup, but replace the GCN layer with feature transformation layers (totaling four such layers). Converting GCN layer to feature transformation layer is equivalent to setting $A=I$ in in GCN layers. We also construct a faster GFN, namely ``GFN-light'', that contains only a single feature transformation layer, which can further reduce the training time while maintaining similar performance.

For both our GCN and GFN, we utilize ReLU activation and batch normalization~\cite{ioffe2015batch}, and fix the hidden dimensionality to 128. No regularization is applied. Furthermore we use batch size of 128, a fixed learning rate of 0.001, and the Adam optimizer~\cite{kingma2014adam}. GLN follows the same setting as GFN, but contains no feature transform layer. It only has the global sum pooling of graph features followed by a single fully connected layer. To compare with existing work, we follow~\cite{xinyi2018capsule,xu2018how} and perform 10-fold cross validation. We run the model for 100 epochs, and select the epoch in the same way as~\cite{xu2018how}, i.e., a single epoch with the best cross-validation accuracy averaged over the 10 folds is selected. We report the average and standard deviation of test accuracies at the selected epoch over 10 folds.

In terms of input node features for GFN and GLN, by default, we use both degree and multi-scale propagated features (up to $K=3$), that is $[\bm d, X, \tilde{A}^1X, \tilde{A}^2X, \tilde{A}^3X]$. We turn discrete features into one-hot vectors, and also discretize degree features into one-hot vectors, as suggested in~\cite{Fey/Lenssen/2019}. We set $X=\vec{1}$ for the social graphs we consider as there are no node attributes. By default, we also augment node features in our GCN with an extra node degree feature (to counter that the normalized adjacency matrix may lose the degree information). Other graph augmented features are also studied for GCN (which has minor effects). All experiments are run on Nvidia GTX 1080 Ti GPU.

\subsection{GFN performs similarly or better compared to its GNN variants across 10 datasets}

Table \ref{tab:methods_bio} and \ref{tab:methods_social} show the results of different methods in both biological and social datasets. It is worth noting that in both datasets, GFN achieves similar performances with our GCN, and match or exceed existing state-of-the-art results on multiple datasets, while GLN performs worse in most of the datasets. This result suggests the importance of non-linear set function, while casting doubt on the necessity of non-linear graph filtering for these benchmarks. Not only that GFN matches GCN performance, but it also performs comparably, if not better, than existing approaches on most of the datsets tested.

\begin{figure*}[!t]
    \centering
    \begin{subfigure}[b]{0.23\textwidth}
        \includegraphics[width=\textwidth]{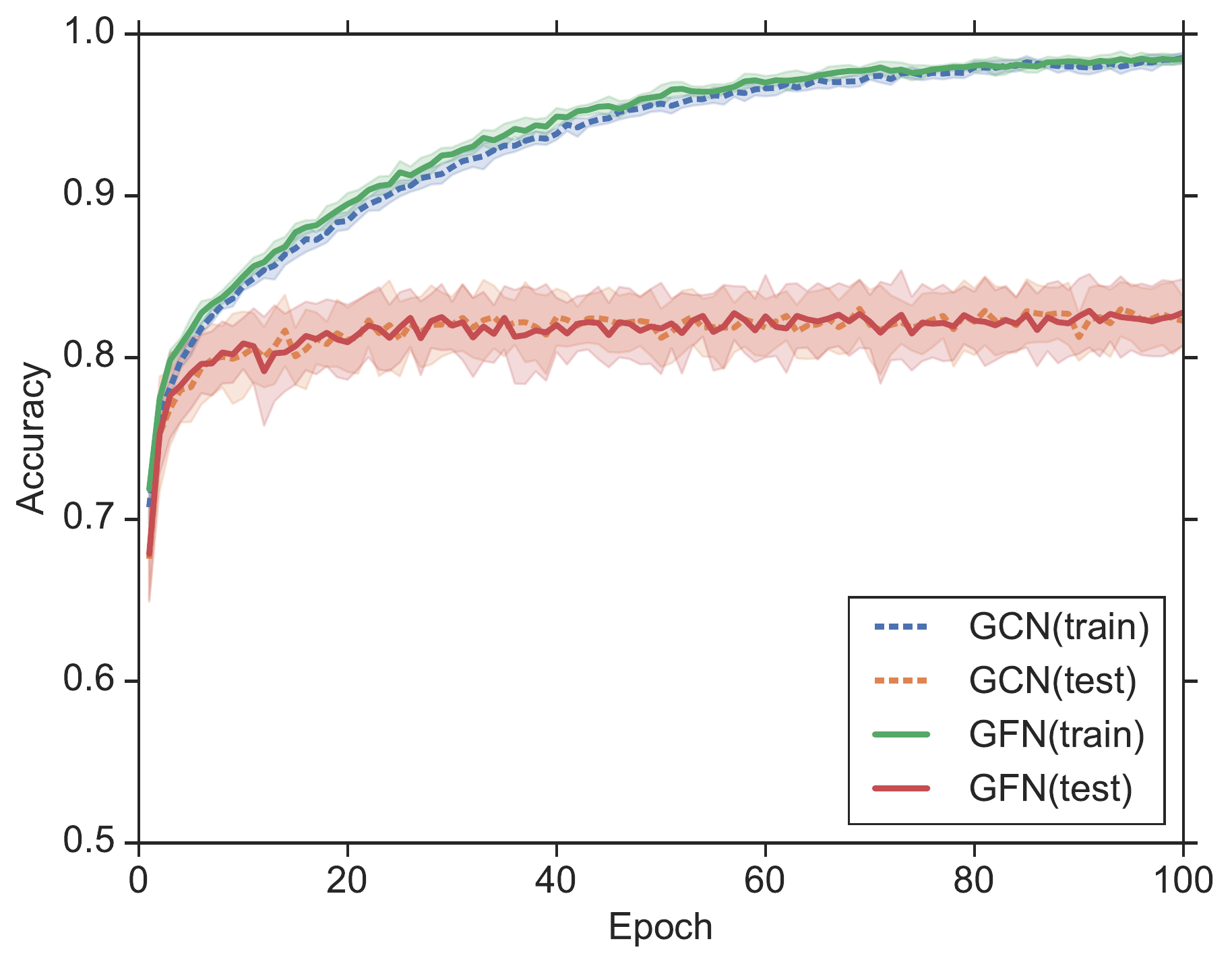}
        \caption{NCI1}
    \end{subfigure}
    \begin{subfigure}[b]{0.23\textwidth}
        \includegraphics[width=\textwidth]{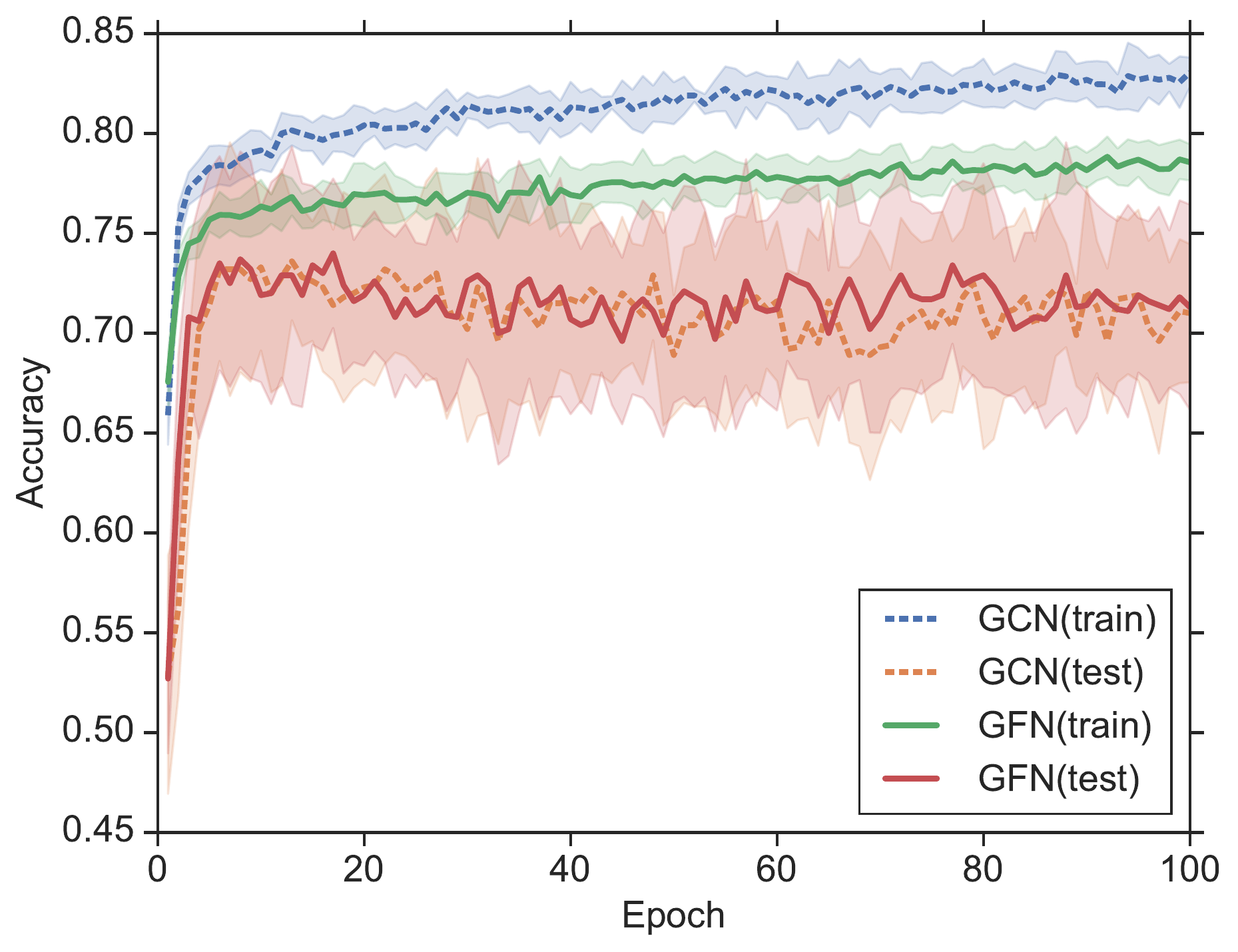}
        \caption{IMDB-BINARY}
    \end{subfigure}
    \begin{subfigure}[b]{0.23\textwidth}
        \includegraphics[width=\textwidth]{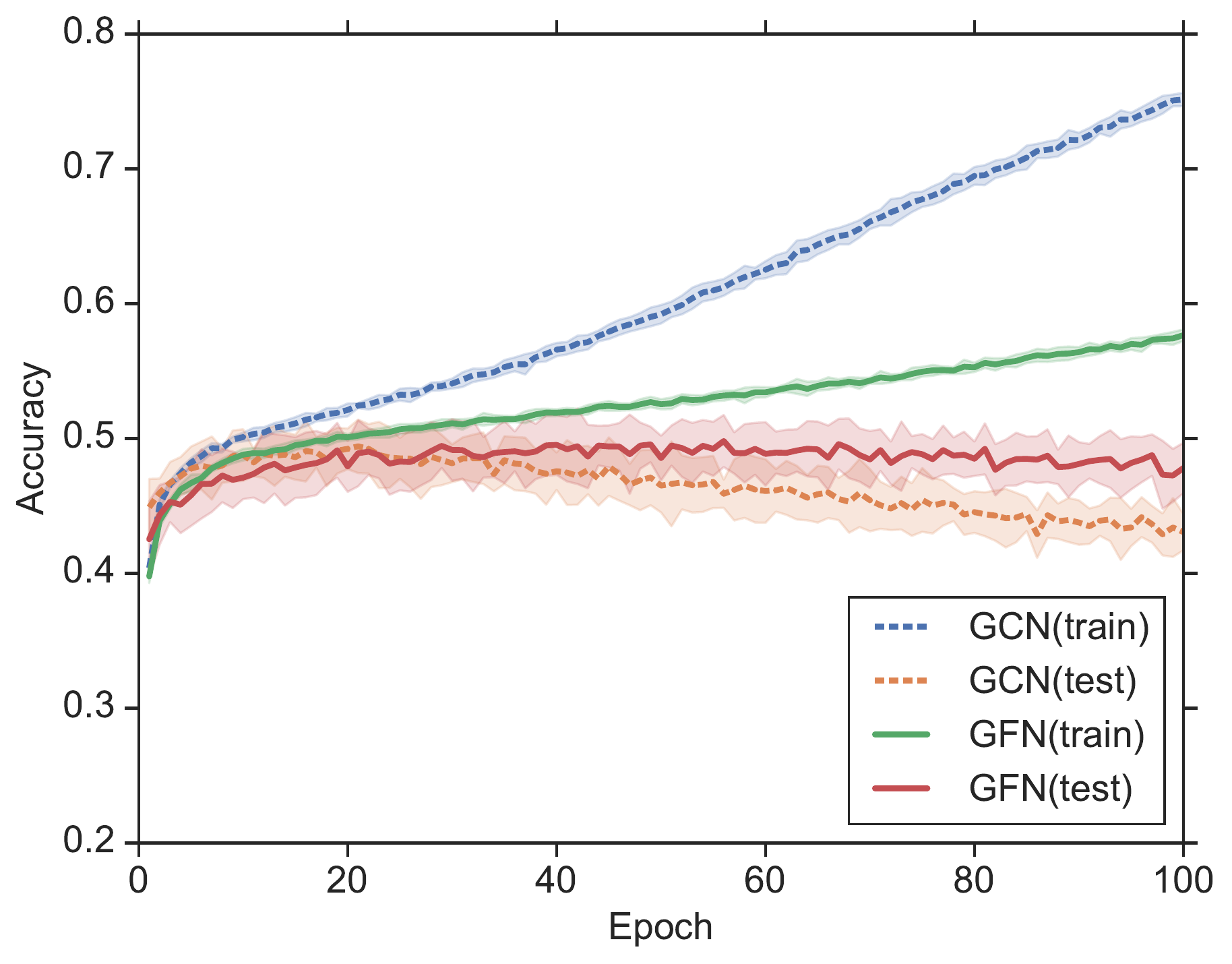}
        \caption{REDDIT-MULTI-12K}
    \end{subfigure}
    \begin{subfigure}[b]{0.23\textwidth}
        \includegraphics[width=\textwidth]{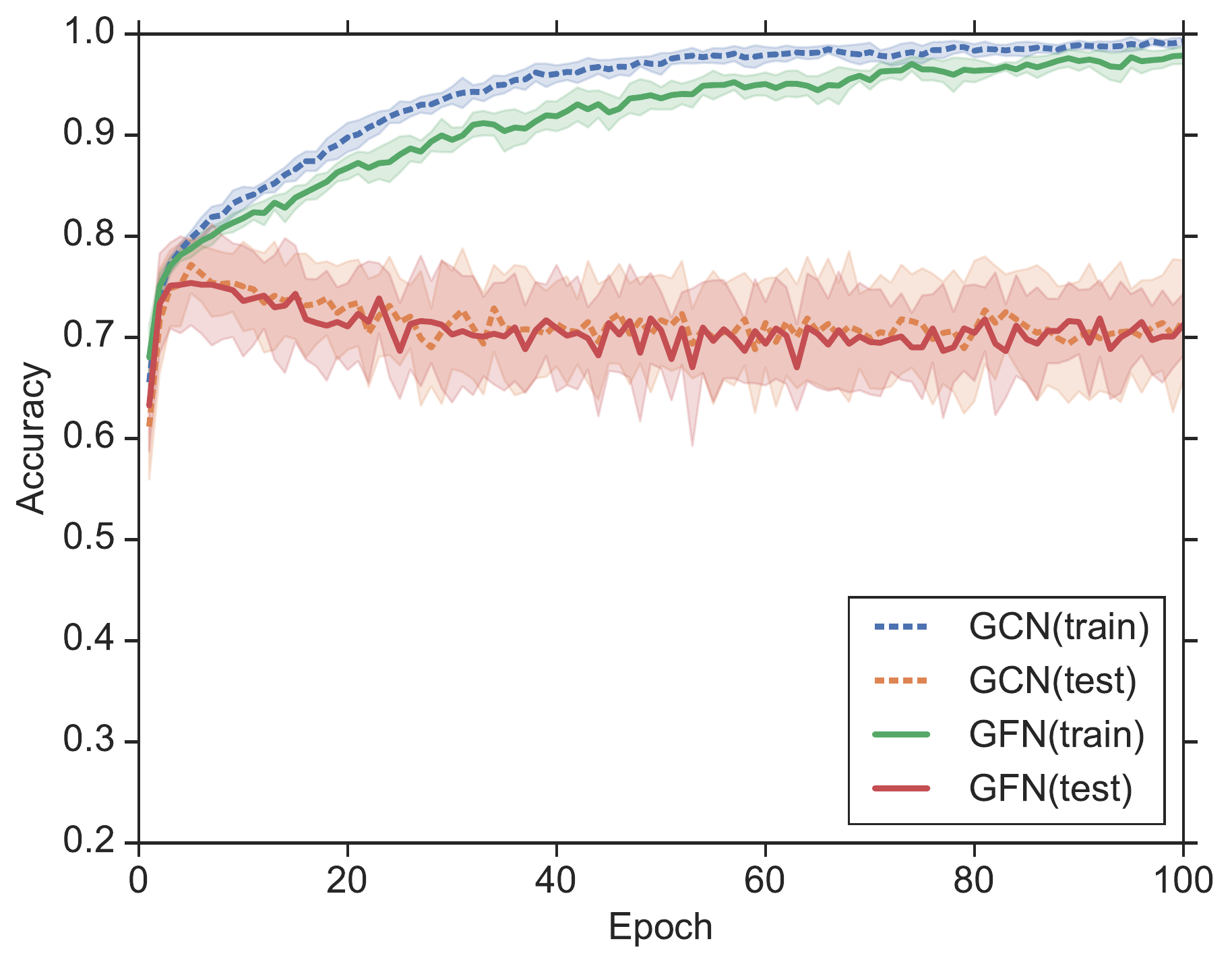}
        \caption{PROTEINS}
    \end{subfigure}
    \caption{\label{fig:perf_vs_epoch} Training and test performance versus training epoch.}
\vspace{-1em}
\end{figure*}

\subsection{While GCN trains better, GFN generalizes better}

To better understand the training dynamic, we plot the training/test curves for both GCN and GFN in Figure \ref{fig:perf_vs_epoch}. We observe that GCN usually performs better than GFN during the training, but GFN can generalize similarly or better than GCN in the test set. This shows that GFN works well not because it is easier to optimize, but its generalization capability. This observation also indicates that linear graph filtering, as used in GFN, may be a good inductive bias for the tested graph classification datasets.

Since GCN overfits, one may wonder if increasing data size could make a difference. To test the impact of dataset size, we take the largest graph dataset available in the benchmark, RE-M12K, which has 11929 graphs. And we then construct nine new datasets by randomly sampling the original dataset with different ratios, ranging from 10\% to 100\% of all graphs. We compute both training and test accuracies over 10 fold cross-validation for both our GFN and GCN. For each dataset size (10 fold cross validation), we consider two ways to extract performance: (1) selecting best epoch averaged over 10 fold cross validation, or (2) selecting the last epoch (i.e. the 100-th epoch). 

\begin{wrapfigure}{r}{0.5\textwidth}
\begin{center}
    \vspace{-1em}
    \centering
    \begin{subfigure}[b]{0.24\textwidth}
        \includegraphics[width=\textwidth]{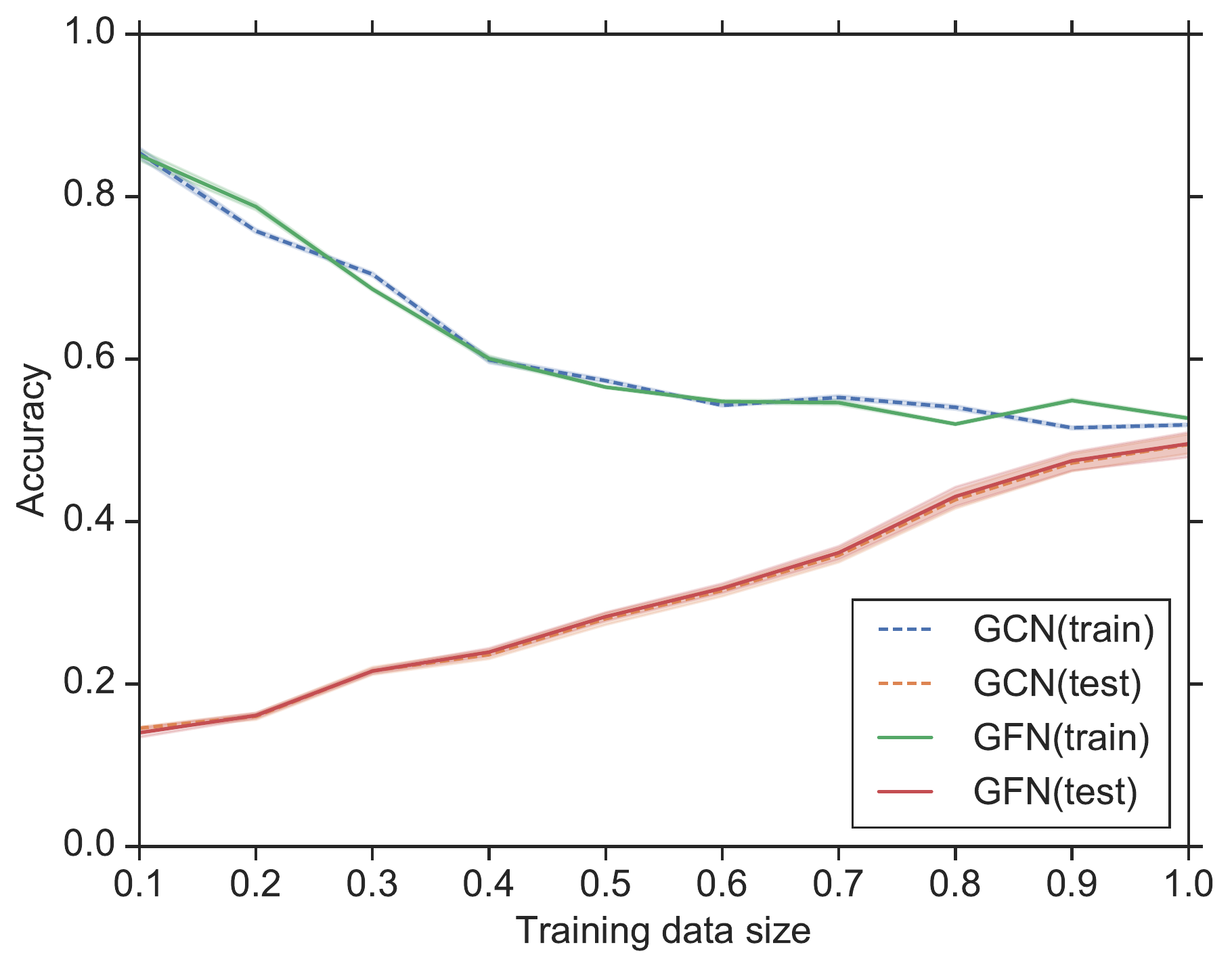}
        \caption{Best epoch}
    \end{subfigure}
    \begin{subfigure}[b]{0.24\textwidth}
        \includegraphics[width=\textwidth]{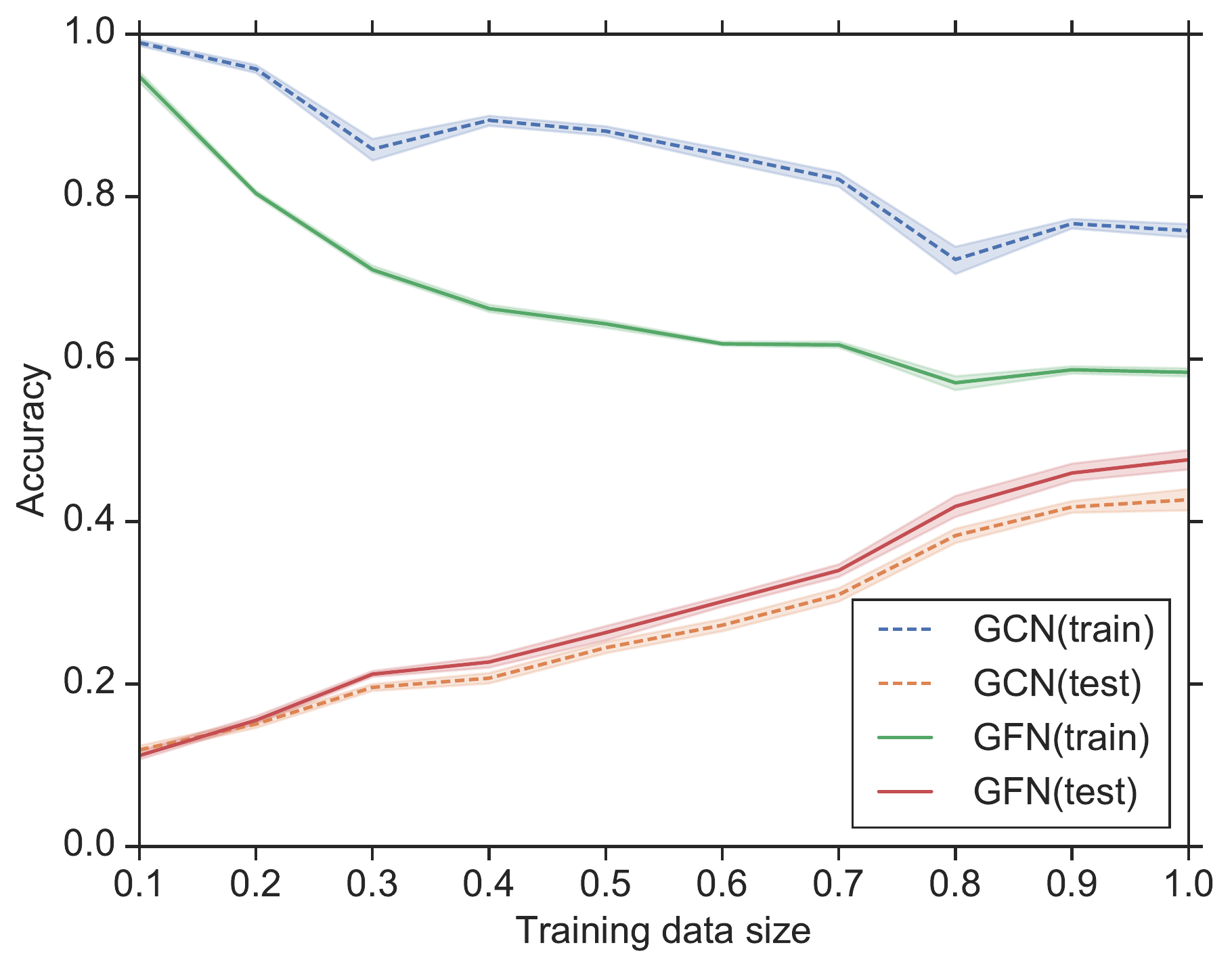}
        \caption{Last epoch.}
    \end{subfigure}
    \caption{\label{fig:perf_vs_percent}Performances on subsets of RE-M12K with different number of training graphs. We see no benefits of increased dataset size for GCN.}
    \vspace{-1em}
\end{center}
\end{wrapfigure}
Figure \ref{fig:perf_vs_percent} shows the results. We can see that as data size increases, 1) it is harder for both models to overfit (training accuracy decreases), but it seems GCN still overfits more if trained longer (to the last epoch); 2) at the best epoch, both models performance almost identical, no sign of differences as training data size increases.

Through these experiments, we conclude that GFN generalizes similarly or better than GCN, and this may be related to the good inductive bias of linear graph filtering, or the inadequacy of the benchmark datasets in testing powerful GNNs. 

\subsection{GFNs are more efficient than GCNs}

\begin{figure*}[!t]
    \centering
    \includegraphics[trim=0 10 0 0,clip,width=\textwidth]{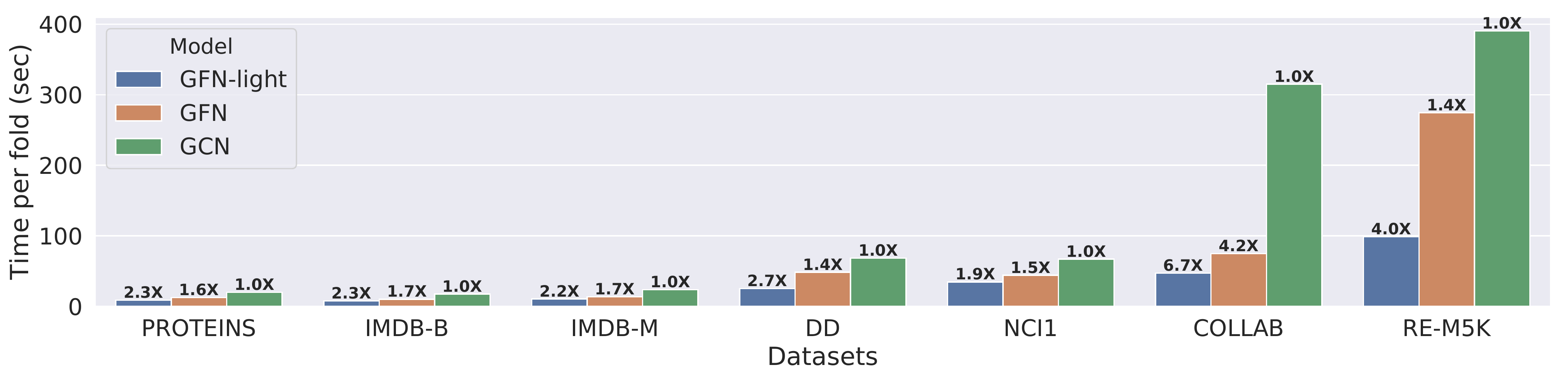}
    \caption{\label{fig:train_time}Training time comparisons. The annotation, e.g. $1.0\times$, denotes speedup compared to GCN.}
    \vspace{-1em}
\end{figure*}

Since GFN's performance is on par with GCN's, we further compare the training time of our GCN and the proposed GFNs. Figure \ref{fig:train_time} shows that a significant speedup (from 1.4$\times$ to $6.7\times$ as fast) by utilizing GFN compared to GCN, especially for datasets with denser edges such as the COLLAB dataset. Also since our GFN can work with fewer transformation layers, GFN-light can achieve better speedup by reducing the number of transformation layers. Note that our GCN is already very efficient as it is built on a highly optimized framework~\cite{Fey/Lenssen/2019}.

\subsection{Ablations}

\begin{table*}[!t]
\footnotesize
\caption{\label{tab:features}Accuracies (\%) under various augmented features. Averaged results over multiple datasets are shown here. $A^{1,2,3}X$ is abbreviated for $A^1X,A^2X,A^3X$, and default node feature $X$ is always used (if available) but not displayed to reduce clutter. Best results per row/block are highlighted.}
\centering
\begin{tabular}{ll|c|cccc|ccc}
\Xhline{1.6\arrayrulewidth}\toprule
Graphs       & Model &   None &  $\bm d$  &  $A^1X$ &   $A^{1,2}X$ &   $A^{1,2,3}X$ &  $\bm d, A^1X$ &   $\bm d, A^{1,2}X$ & $\bm d,A^{1,2,3}X$ \\
\midrule\midrule
\multirow{2}{*}{Bio.} & GCN &  \textbf{78.52} &     78.51 &  78.23 &  78.24 &  \textbf{78.68} &         79.10 &         79.26 &         \textbf{79.69} \\
       & GFN &  \textbf{76.27} &     77.84 &  78.78 &  79.09 &  \textbf{79.17} &         78.71 &        \textbf{ 79.21 }&         79.13 \\
\midrule
\multirow{2}{*}{Soical} & GCN &  \textbf{34.02} &   \textbf{  62.35} &  59.20 &  60.39 &  60.28 &         62.45 &         62.71 &        \textbf{ 62.77 }\\
       & GFN & \textbf{30.45} &    \textbf{ 60.79} &  58.04 &  59.83 &  60.09 &         62.47 &         \textbf{62.63 }&         62.60 \\
\bottomrule\Xhline{1.6\arrayrulewidth}
\end{tabular}
\vspace{-1em}
\end{table*}

\textbf{Node features.} To better understand the impact of features, we test both models with different input node features. Table \ref{tab:features} shows that 1) graph features are very important for both GFN and GCN, 2) the node degree feature is surprisingly important, and multi-scale features can further improve on that, and 3) even with multi-scale features, GCN still performs similarly to GFN, which further suggests that linear graph filtering is enough. More detailed results (per dataset) can be found in the \wappendix.

\begin{wraptable}{r}{0.5\textwidth}
\vspace{-1em}
\tiny
\centering
\caption{\label{tab:layers}Accuracies (\%) under different number of Conv. layers. GFN requires fewer layers to achieve similar results as GCN.}
\begin{tabular}{ll|ccccc}
\Xhline{1.6\arrayrulewidth}\toprule
       &  &     1 &      2 &      3 &      4 &      5\\
\midrule\midrule
\multirow{2}{*}{Bio.} & GCN & 77.17 &  \textbf{79.38} &  78.86 &  78.75 &  78.21 \\
       & GFN & 79.59 &  79.77 &  \textbf{79.78} &  78.99 &  78.14 \\
\midrule
\multirow{2}{*}{Soical} & GCN & 60.69 &  62.12 &  62.37 &  \textbf{62.70} &  62.46 \\
       & GFN &  62.70 &  \textbf{62.88} &  62.81 &  62.80 &  62.60 \\
\bottomrule\Xhline{1.6\arrayrulewidth}
\end{tabular}
\vspace{-1em}
\end{wraptable}
\textbf{Architecture depth.} We vary the number of convolutional layers (with two FC-layers after sum pooling kept the same). Table \ref{tab:layers} shows that 1) GCN benefits from multiple grpah convolutional layers with a significant diminishing return, 2) GFN with single feature transformation layer works pretty well already, likely due to the availability of multi-scale input node features, which otherwise require multiple GCN layers to obtain.

 \vspace{-1em}
\section{Related Work}

Our work studies the effectiveness of GNNs via linearization, and propose a simple, effective and efficient architecture. Two family of work are most relevant to ours. 

Many new graph neural networks architectures (for graph classification) have been proposed recently, such as~\cite{kipf2016semi,hamilton2017inductive,zhang2018end,xu2018representation,liao2019lanczosnet,gao2019graph,maron2019provably}. Many of these networks aims to improve the expressiveness or trainability of GNNs. We do not aim to propose a more powerful architecture, our GFN aims for simplicity and efficiency, at the same time can obtain high performance on graph classification benchmark that is on-par with state-of-the-art.

There have been relatively fewer work on understanding graph neural network. Existing work mainly focus on studying the expressiveness of GNNs~\cite{xu2018how,nt2019revisiting,dehmamy2019understanding}, in theoretical limit. We take a more practical route by examining the effects of linearization on real graph classification problems, leading to a scalable approach (i.e. GFN) that works well.
 \vspace{-1em}\section{Discussion}
\label{sec:discuss}
In this work, we conduct a dissection of GNNs on common graph classification benchmarks. We first decompose GNNs into two parts, and linearize the graph filtering part resulting GFN. We then further linearize the set function of GFN resulting GLN. In our extensive experiments, we find GFN can match or exceed the best results by recently proposed GNNs, with a fraction of computation cost. The linearization of graph filtering (i.e. GFN) has little impact on performance, while linearization of both graph filtering and set function (i.e. GLN) leads to worse performance. 

Since GCN usually achieve better training accuracies while not better test accuracies, we conjecture that the linear graph filtering may be a good inductive bias for tested datasets, though this is speculative and requires more future investigations. It also casts doubts on the adequacy of existing graph classification benchmarks. It is possible that the complexity of current graph classification benchmarks is limited, so that linear graph filtering is enough, and moving to datasets or problems with higher structural complexity could require sophisticated non-linear graph filtering.

\section*{Acknowledgements}
We would like to thank Yunsheng Bai and Zifeng Kang for their help in a related project prior to this work. We also thank Jascha Sohl-dickstein, Jiaxuan You, Kevin Swersky, Yasaman Bahri, Yewen Wang, Ziniu Hu and Allan Zhou for helpful discussions and feedbacks. This work is partially supported by NSF III-1705169, NSF CAREER Award 1741634, and Amazon Research Award.

{\small
\bibliography{main}

\begin{thebibliography}{31}
\providecommand{\natexlab}[1]{#1}
\providecommand{\url}[1]{\texttt{#1}}
\expandafter\ifx\csname urlstyle\endcsname\relax
  \providecommand{\doi}[1]{doi: #1}\else
  \providecommand{\doi}{doi: \begingroup \urlstyle{rm}\Url}\fi

\bibitem[Scarselli et~al.(2009)Scarselli, Gori, Tsoi, Hagenbuchner, and
  Monfardini]{scarselli2009graph}
Franco Scarselli, Marco Gori, Ah~Chung Tsoi, Markus Hagenbuchner, and Gabriele
  Monfardini.
\newblock The graph neural network model.
\newblock \emph{IEEE Transactions on Neural Networks}, 2009.

\bibitem[Li et~al.(2015)Li, Tarlow, Brockschmidt, and Zemel]{li2015gated}
Yujia Li, Daniel Tarlow, Marc Brockschmidt, and Richard Zemel.
\newblock Gated graph sequence neural networks.
\newblock \emph{arXiv preprint arXiv:1511.05493}, 2015.

\bibitem[Defferrard et~al.(2016)Defferrard, Bresson, and
  Vandergheynst]{defferrard2016convolutional}
Micha{\"e}l Defferrard, Xavier Bresson, and Pierre Vandergheynst.
\newblock Convolutional neural networks on graphs with fast localized spectral
  filtering.
\newblock In \emph{Advances in neural information processing systems}, 2016.

\bibitem[Kipf and Welling(2016)]{kipf2016semi}
Thomas~N Kipf and Max Welling.
\newblock Semi-supervised classification with graph convolutional networks.
\newblock \emph{arXiv preprint arXiv:1609.02907}, 2016.

\bibitem[Wu et~al.(2019)Wu, Zhang, Souza~Jr, Fifty, Yu, and
  Weinberger]{wu2019simplifying}
Felix Wu, Tianyi Zhang, Amauri Holanda~de Souza~Jr, Christopher Fifty, Tao Yu,
  and Kilian~Q Weinberger.
\newblock Simplifying graph convolutional networks.
\newblock \emph{arXiv preprint arXiv:1902.07153}, 2019.

\bibitem[Simonovsky and Komodakis(2017)]{simonovsky2017dynamic}
Martin Simonovsky and Nikos Komodakis.
\newblock Dynamic edge-conditioned filters in convolutional neural networks on
  graphs.
\newblock In \emph{Proceedings of the IEEE Conference on Computer Vision and
  Pattern Recognition}, 2017.

\bibitem[Xinyi and Chen(2019)]{xinyi2018capsule}
Zhang Xinyi and Lihui Chen.
\newblock Capsule graph neural network.
\newblock In \emph{International Conference on Learning Representations}, 2019.
\newblock URL \url{https://openreview.net/forum?id=Byl8BnRcYm}.

\bibitem[Zeiler and Fergus(2014)]{zeiler2014visualizing}
Matthew~D Zeiler and Rob Fergus.
\newblock Visualizing and understanding convolutional networks.
\newblock In \emph{European conference on computer vision}, 2014.

\bibitem[He et~al.(2016)He, Zhang, Ren, and Sun]{he2016identity}
Kaiming He, Xiangyu Zhang, Shaoqing Ren, and Jian Sun.
\newblock Identity mappings in deep residual networks.
\newblock In \emph{European conference on computer vision}, 2016.

\bibitem[Yanardag and Vishwanathan(2015)]{yanardag2015deep}
Pinar Yanardag and SVN Vishwanathan.
\newblock Deep graph kernels.
\newblock In \emph{Proceedings of the 21th ACM SIGKDD International Conference
  on Knowledge Discovery and Data Mining}, 2015.

\bibitem[Zhang et~al.(2018{\natexlab{a}})Zhang, Cui, Neumann, and
  Chen]{zhang2018end}
Muhan Zhang, Zhicheng Cui, Marion Neumann, and Yixin Chen.
\newblock An end-to-end deep learning architecture for graph classification.
\newblock In \emph{Thirty-Second AAAI Conference on Artificial Intelligence},
  2018{\natexlab{a}}.

\bibitem[Dai et~al.(2016)Dai, Dai, and Song]{dai2016discriminative}
Hanjun Dai, Bo~Dai, and Le~Song.
\newblock Discriminative embeddings of latent variable models for structured
  data.
\newblock In \emph{International conference on machine learning}, 2016.

\bibitem[Gilmer et~al.(2017)Gilmer, Schoenholz, Riley, Vinyals, and
  Dahl]{gilmer2017neural}
Justin Gilmer, Samuel~S Schoenholz, Patrick~F Riley, Oriol Vinyals, and
  George~E Dahl.
\newblock Neural message passing for quantum chemistry.
\newblock In \emph{Proceedings of the 34th International Conference on Machine
  Learning}, 2017.

\bibitem[Hamilton et~al.(2017)Hamilton, Ying, and
  Leskovec]{hamilton2017inductive}
Will Hamilton, Zhitao Ying, and Jure Leskovec.
\newblock Inductive representation learning on large graphs.
\newblock In \emph{Advances in Neural Information Processing Systems}, 2017.

\bibitem[Xu et~al.(2019)Xu, Hu, Leskovec, and Jegelka]{xu2018how}
Keyulu Xu, Weihua Hu, Jure Leskovec, and Stefanie Jegelka.
\newblock How powerful are graph neural networks?
\newblock In \emph{International Conference on Learning Representations}, 2019.
\newblock URL \url{https://openreview.net/forum?id=ryGs6iA5Km}.

\bibitem[Klicpera et~al.(2019)Klicpera, Bojchevski, and
  Günnemann]{klicpera2018combining}
Johannes Klicpera, Aleksandar Bojchevski, and Stephan Günnemann.
\newblock Combining neural networks with personalized pagerank for
  classification on graphs.
\newblock In \emph{International Conference on Learning Representations}, 2019.
\newblock URL \url{https://openreview.net/forum?id=H1gL-2A9Ym}.

\bibitem[Zaheer et~al.(2017)Zaheer, Kottur, Ravanbakhsh, Poczos, Salakhutdinov,
  and Smola]{zaheer2017deep}
Manzil Zaheer, Satwik Kottur, Siamak Ravanbakhsh, Barnabas Poczos, Ruslan~R
  Salakhutdinov, and Alexander~J Smola.
\newblock Deep sets.
\newblock In \emph{Advances in neural information processing systems}, 2017.

\bibitem[Shervashidze et~al.(2011)Shervashidze, Schweitzer, Leeuwen, Mehlhorn,
  and Borgwardt]{shervashidze2011weisfeiler}
Nino Shervashidze, Pascal Schweitzer, Erik Jan~van Leeuwen, Kurt Mehlhorn, and
  Karsten~M Borgwardt.
\newblock Weisfeiler-lehman graph kernels.
\newblock \emph{Journal of Machine Learning Research}, 2011.

\bibitem[Ivanov and Burnaev(2018)]{ivanov2018anonymous}
Sergey Ivanov and Evgeny Burnaev.
\newblock Anonymous walk embeddings.
\newblock \emph{arXiv preprint arXiv:1805.11921}, 2018.

\bibitem[Zhang et~al.(2018{\natexlab{b}})Zhang, Wang, Xiang, Huang, and
  Nehorai]{zhang2018retgk}
Zhen Zhang, Mianzhi Wang, Yijian Xiang, Yan Huang, and Arye Nehorai.
\newblock Retgk: Graph kernels based on return probabilities of random walks.
\newblock In \emph{Advances in Neural Information Processing Systems}, pages
  3964--3974, 2018{\natexlab{b}}.

\bibitem[Du et~al.(2019)Du, Hou, P{\'o}czos, Salakhutdinov, Wang, and
  Xu]{Du2019GraphNT}
Simon~S. Du, Kangcheng Hou, Barnab{\'a}s P{\'o}czos, Ruslan Salakhutdinov,
  Ruosong Wang, and Keyulu Xu.
\newblock Graph neural tangent kernel: Fusing graph neural networks with graph
  kernels.
\newblock \emph{ArXiv}, abs/1905.13192, 2019.

\bibitem[Niepert et~al.(2016)Niepert, Ahmed, and Kutzkov]{niepert2016learning}
Mathias Niepert, Mohamed Ahmed, and Konstantin Kutzkov.
\newblock Learning convolutional neural networks for graphs.
\newblock In \emph{International conference on machine learning}, 2016.

\bibitem[Ioffe and Szegedy(2015)]{ioffe2015batch}
Sergey Ioffe and Christian Szegedy.
\newblock Batch normalization: Accelerating deep network training by reducing
  internal covariate shift.
\newblock \emph{arXiv preprint arXiv:1502.03167}, 2015.

\bibitem[Kingma and Ba(2014)]{kingma2014adam}
Diederik~P Kingma and Jimmy Ba.
\newblock Adam: A method for stochastic optimization.
\newblock \emph{arXiv preprint arXiv:1412.6980}, 2014.

\bibitem[Fey and Lenssen(2019)]{Fey/Lenssen/2019}
Matthias Fey and Jan~E. Lenssen.
\newblock Fast graph representation learning with {PyTorch Geometric}.
\newblock In \emph{ICLR Workshop on Representation Learning on Graphs and
  Manifolds}, 2019.

\bibitem[Xu et~al.(2018)Xu, Li, Tian, Sonobe, Kawarabayashi, and
  Jegelka]{xu2018representation}
Keyulu Xu, Chengtao Li, Yonglong Tian, Tomohiro Sonobe, Ken-ichi Kawarabayashi,
  and Stefanie Jegelka.
\newblock Representation learning on graphs with jumping knowledge networks.
\newblock \emph{arXiv preprint arXiv:1806.03536}, 2018.

\bibitem[Liao et~al.(2019)Liao, Zhao, Urtasun, and Zemel]{liao2019lanczosnet}
Renjie Liao, Zhizhen Zhao, Raquel Urtasun, and Richard~S Zemel.
\newblock Lanczosnet: Multi-scale deep graph convolutional networks.
\newblock \emph{arXiv preprint arXiv:1901.01484}, 2019.

\bibitem[Gao and Ji(2019)]{gao2019graph}
Hongyang Gao and Shuiwang Ji.
\newblock Graph u-nets.
\newblock \emph{arXiv preprint arXiv:1905.05178}, 2019.

\bibitem[Maron et~al.(2019)Maron, Ben-Hamu, Serviansky, and
  Lipman]{maron2019provably}
Haggai Maron, Heli Ben-Hamu, Hadar Serviansky, and Yaron Lipman.
\newblock Provably powerful graph networks.
\newblock In \emph{Advances in Neural Information Processing Systems}, pages
  2153--2164, 2019.

\bibitem[NT and Maehara(2019)]{nt2019revisiting}
Hoang NT and Takanori Maehara.
\newblock Revisiting graph neural networks: All we have is low-pass filters.
\newblock \emph{arXiv preprint arXiv:1905.09550}, 2019.

\bibitem[Dehmamy et~al.(2019)Dehmamy, Barab{\'a}si, and
  Yu]{dehmamy2019understanding}
Nima Dehmamy, Albert-L{\'a}szl{\'o} Barab{\'a}si, and Rose Yu.
\newblock Understanding the representation power of graph neural networks in
  learning graph topology.
\newblock In \emph{Advances in Neural Information Processing Systems}, pages
  15387--15397, 2019.

\end{thebibliography}
\bibliographystyle{unsrtnat}}

\newpage
\appendix

\section{Comparisons of different linearizations}

Table \ref{tab:linear_sum} summarizes the comparisons between GCN and its linearized variants. The efficiency and performance are concluded from our experiments on graph classification benchmarks. Noted that a GNN with a linear graph filtering can be seen as a GFN, but the reverse may not be true. General GFN can have non-linear graph filtering, e.g. when the feature extraction function $\gamma(G, X)$ is not a linear map of $X$. Thus we use GFN$^{lin}$ in Table \ref{tab:linear_sum} to denote such subtle difference.

\begin{table*}[h]
\begin{center}
\begin{small}
\caption{\label{tab:linear_sum}Comparisons of different linearizations for GNN/GCN. Here we denote the GFN as GFN$^{lin}$ since the GFN we study in this work has linear graph filtering, but general GFN can also have a non-linear graph filtering function.}
\begin{tabular}{lllll}
\toprule
\bf Method &\bf Graph filtering &\bf Set function  &\bf Efficiency &\bf Performance
\\ \midrule
GLN & Linear & Linear & High & Low \\
GFN$^{lin}$ & Linear & Non-linear & High & High \\
GCN & Non-linear & Linear/Non-linear & Low & High \\
\bottomrule
\end{tabular}
\end{small}
\end{center}
\end{table*}

\section{GFN can be a generic framework for functions over graphs}

Beyond as a tool to study GNN parts, GFN is also more efficient than GNN counterpart, which makes it a fast approximation. Furthermore, GFNs can be a very powerful framework without restriction on the feature extraction function $\gamma(G, X)$ and the exact forms of the set function. The potential expressiveness of a GFN is demonstrated by the following proposition. 

\begin{proposition}
\label{th:gfn_power}
For any GNN $\mathcal{F}$ defined in $\mathcal{G}_X$, there exists a graph to set mapping $\mathcal{M}: \mathcal{G}\rightarrow \mathcal{S}$ where $\mathcal{S}$ is a set space, and a set function $\mathcal{T}$ that approximates $\mathcal{F}$ to arbitrary precision, i.e. $\forall G\in \mathcal{G}_X, F(G) \approx \mathcal{T}(\mathcal{M}(G))$.
\end{proposition}

The proof is provided below. We want to provide an intuitive interpretation here. There exists some way(s) that we can encode any graph into a set, and learn a generic set function on it. As long as the set contains the graph information, a powerful set function can learn to integrate it in a flexible way. So a well constructed GFN can be as powerful as, if not more powerful than, the most powerful GNNs. This shows the potential of the GFN framework in modeling arbitrary graph data.

Here we provide the proof for Proposition \ref{th:gfn_power}.

\begin{proof}
We show the existence of the mapping $\mathcal{T}$ by constructing it as follows. First, we assign a unique ID to each of the node, then we add its ID and its neighbors' IDs in the end of node features. If there are edges with features, we also treat them as nodes and apply the same above procedure. This procedure results in a set of nodes with features that preserve the same original information (since we can reconstruct the original graph).

We now show the existence of a set function that can mimic any graph functions operated on $\mathcal{G}$, again, by constructing a specific one. Since the set of nodes preserve the whole graph information, the set function can first reconstruct the graph by decoding the node's feature vectors. At every computation step, the set function find neighbors of each node in the set, and compute the aggregation function in exactly the same way as the graph function would do with the neighbors of a node. This procedure is repeated until the graph function produces its output.

Hence, the above constructed example proves the existence of $\mathcal{M}$ and a set function $\mathcal{T}$ such that $\forall G\in \mathcal{G}_X, \mathcal{F}(G) \approx \mathcal{T}(\mathcal{M}(G))$. We also note that the specially constructed examples above are feasible but likely not optimal. A better solution is to have a set function that learns to adaptively leverage the graph structure as well as node attributes.
\end{proof}

\section{Proof for proposition 1}

Here we provide the proof for Proposition \ref{th:prop1}.

\begin{proof}
According to claim \ref{th:claim} and definition \ref{th:def3}, a $\text{GNN}(G, X)$ with a linear graph filtering part, denoted by $\text{GNN}^{lin}(G, X)$, can be written as follows.
$$\text{GNN}^{lin}(G, X) = \mathcal{T}\circ \mathcal{F}_G(X)=\mathcal{T}(\Gamma(G, X)\bm\theta)=\mathcal{T}'(\Gamma(G, X)),$$
where $\bm\theta$ is absorbed into the set function $\mathcal{T}'(\cdot)$.
According to GFN's definition in Eq. \ref{eq:gfn_2} and general set function result from~\cite{zaheer2017deep}, we have
$$\text{GFN}(G, X) = \mathcal{T}''(X^G) = \mathcal{T}''(\gamma(G, X)).$$
By setting $\gamma(G, X)=\Gamma(G, X)$, we arrive at $\text{GNN}^{lin}(G, X) = \text{GFN}(G, X)$.
\end{proof} 

\section{Dataset details}

Table \ref{tab:data_stat_bio} data statistics for biological graph, including MUTAG, NCI1, PROTEINS, D\&D, ENZYMES. Table  \ref{tab:data_stat_social} shows data statistics for social graphs, including COLLAB, IMDB-Binary (IMDB-B), IMDB-Multi (IMDB-M), Reddit-Multi-5K (RE-M5K), Reddit-Multi-12K (RE-M12K). It is worth noting that ENZYMES dataset we use has 3 node label + 18 continuous features, and PROTEINS dataset we use has 3 node label plus 1 continuous feature. And we use all available features as input to our GCN, GFN as well as GLN.

\begin{table}[h]
\small
\centering
\caption{\label{tab:data_stat_bio}Data statistics of Biological dataset}
\begin{tabular}{lcccccc}
\Xhline{1.6\arrayrulewidth}\toprule  
Dataset & MUTAG & NCI1 & PROTEINS & D\&D & ENZYMES \\
\midrule\midrule
\# graphs & 188 & 4110 & 1113 & 1178 & 600 \\
\# classes & 2 & 2 & 2 & 2 & 6 \\
\# features & 7 & 37 & 4 & 82 & 21 \\
Avg \# nodes & 17.93 & 29.87 & 39.06 & 284.32 & 32.63 \\
Avg \# edges & 19.79 & 32.30 & 72.82 & 715.66 & 62.14 \\
\bottomrule\Xhline{1.6\arrayrulewidth}
\end{tabular}
\end{table}

\begin{table}[h]
\small
\centering
\caption{\label{tab:data_stat_social}Data statistics of Social dataset}
\begin{tabular}{lccccc}
\Xhline{1.6\arrayrulewidth}\toprule  
Dataset & COLLAB & IMDB-B & IMDB-M & RE-M5K & RE-M12K \\
\midrule\midrule
\# graphs & 5000 & 1000 & 1500 & 4999 & 11929 \\
\# classes & 3 & 2 & 3 & 5 & 11 \\
\# features & 1 & 1 & 1 & 1 & 1 \\
Avg \# nodes & 74.49 & 19.77 & 13.00 & 508.52 & 391.41 \\
Avg \# edges & 2457.78 & 96.53 & 65.94 & 594.87 & 456.89 \\
\bottomrule\Xhline{1.6\arrayrulewidth}
\end{tabular}
\end{table}

\section{Extra curves on training / test performance vs epoch}

\begin{figure}[h]
\centering
    \begin{subfigure}[b]{0.3\textwidth}
        \includegraphics[width=\textwidth]{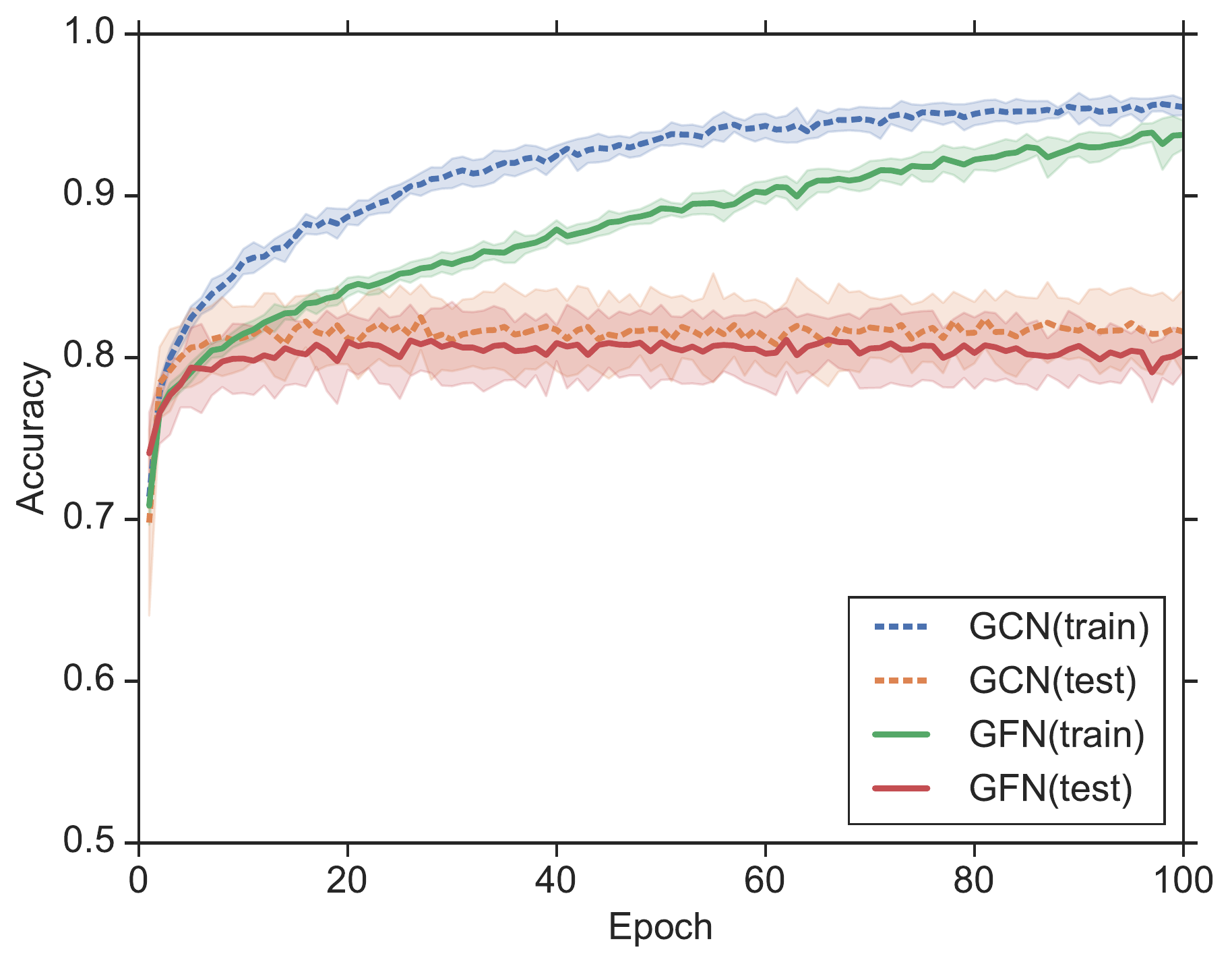}
        \caption{COLLAB}
    \end{subfigure}
    \begin{subfigure}[b]{0.3\textwidth}
        \includegraphics[width=\textwidth]{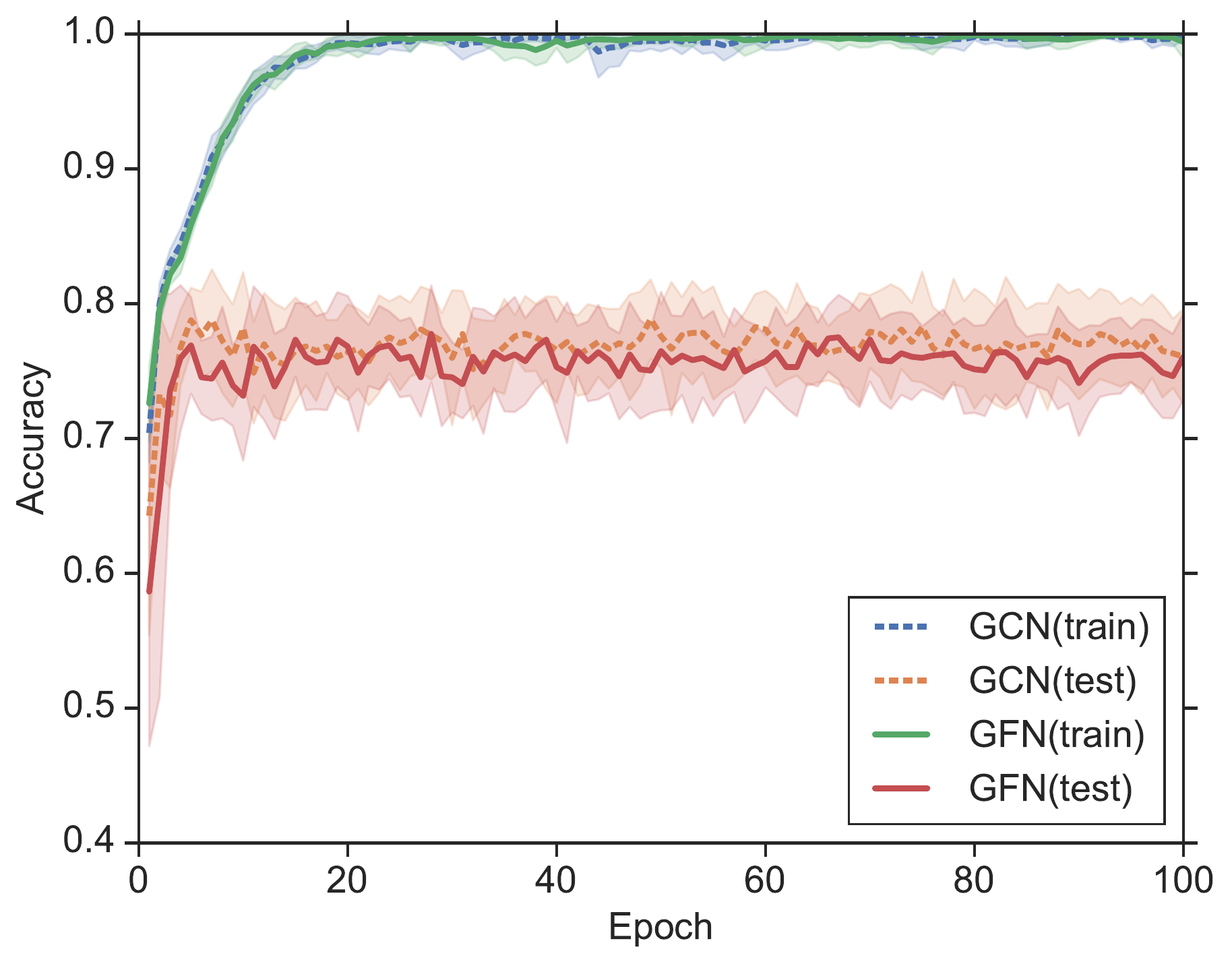}
        \caption{DD}
    \end{subfigure}
    \begin{subfigure}[b]{0.3\textwidth}
        \includegraphics[width=\textwidth]{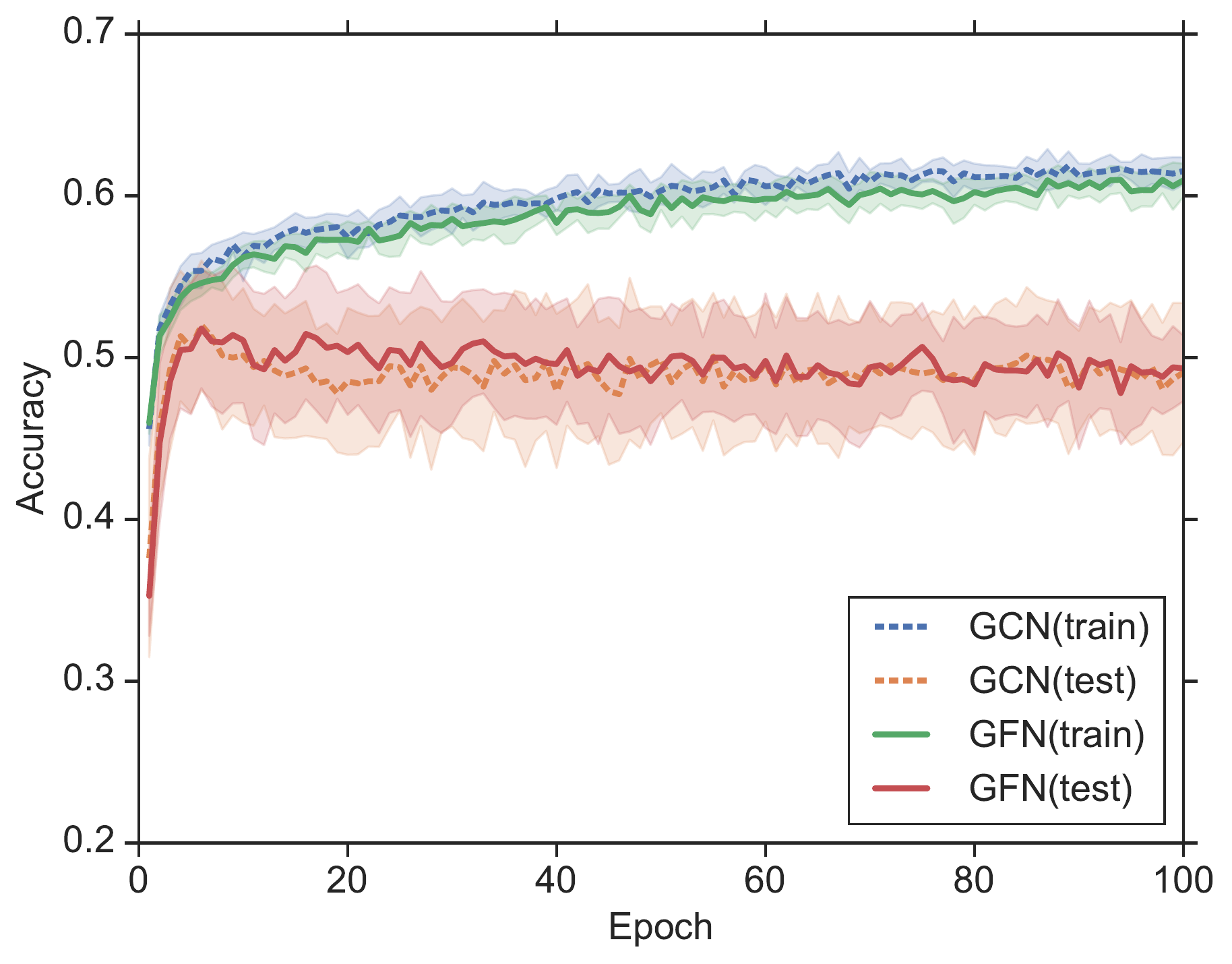}
        \caption{IMDB-MULTI}
    \end{subfigure}
    \begin{subfigure}[b]{0.3\textwidth}
        \includegraphics[width=\textwidth]{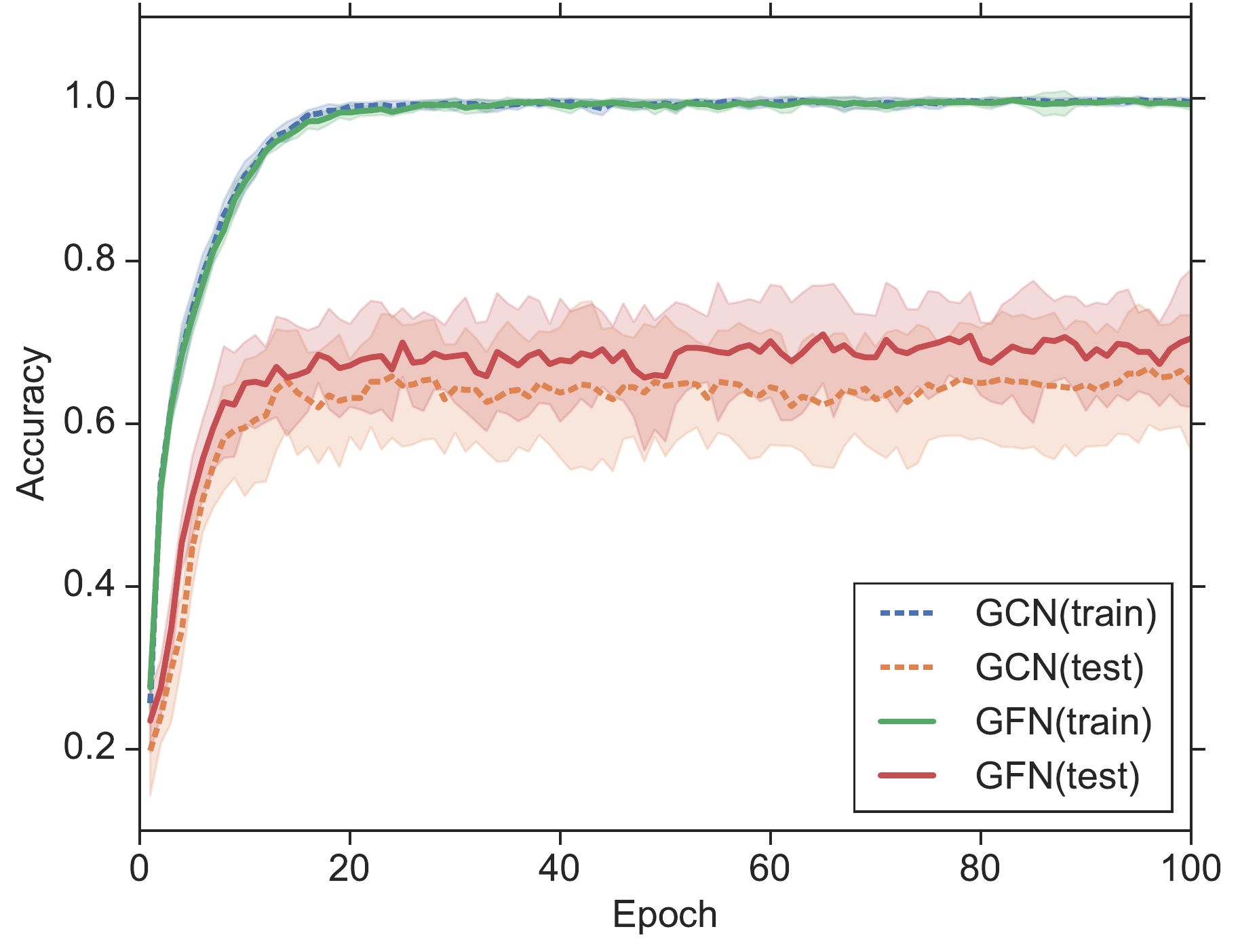}
        \caption{ENZYMES}
    \end{subfigure}
    \begin{subfigure}[b]{0.3\textwidth}
        \includegraphics[width=\textwidth]{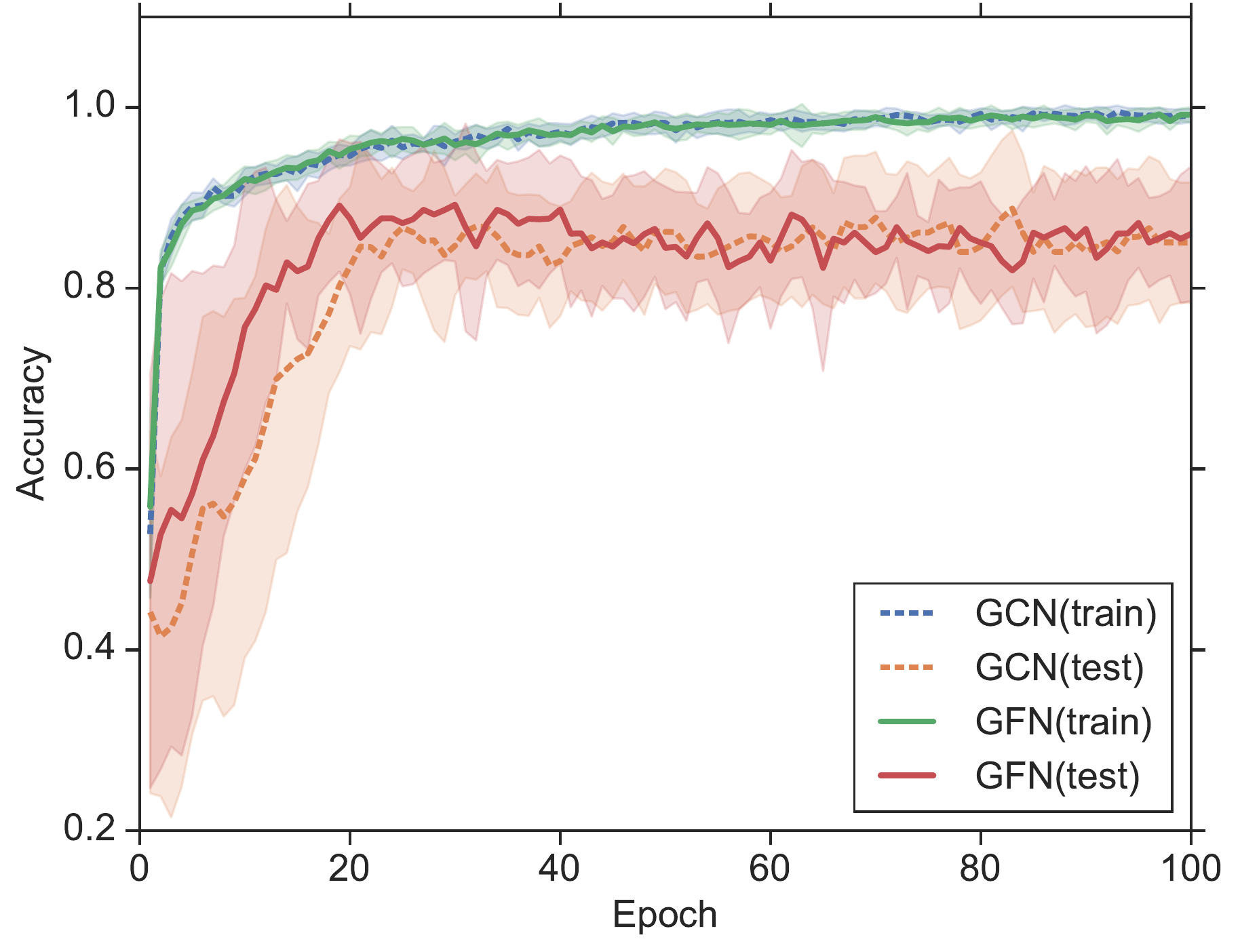}
        \caption{MUTAG}
    \end{subfigure}
    \begin{subfigure}[b]{0.3\textwidth}
        \includegraphics[width=\textwidth]{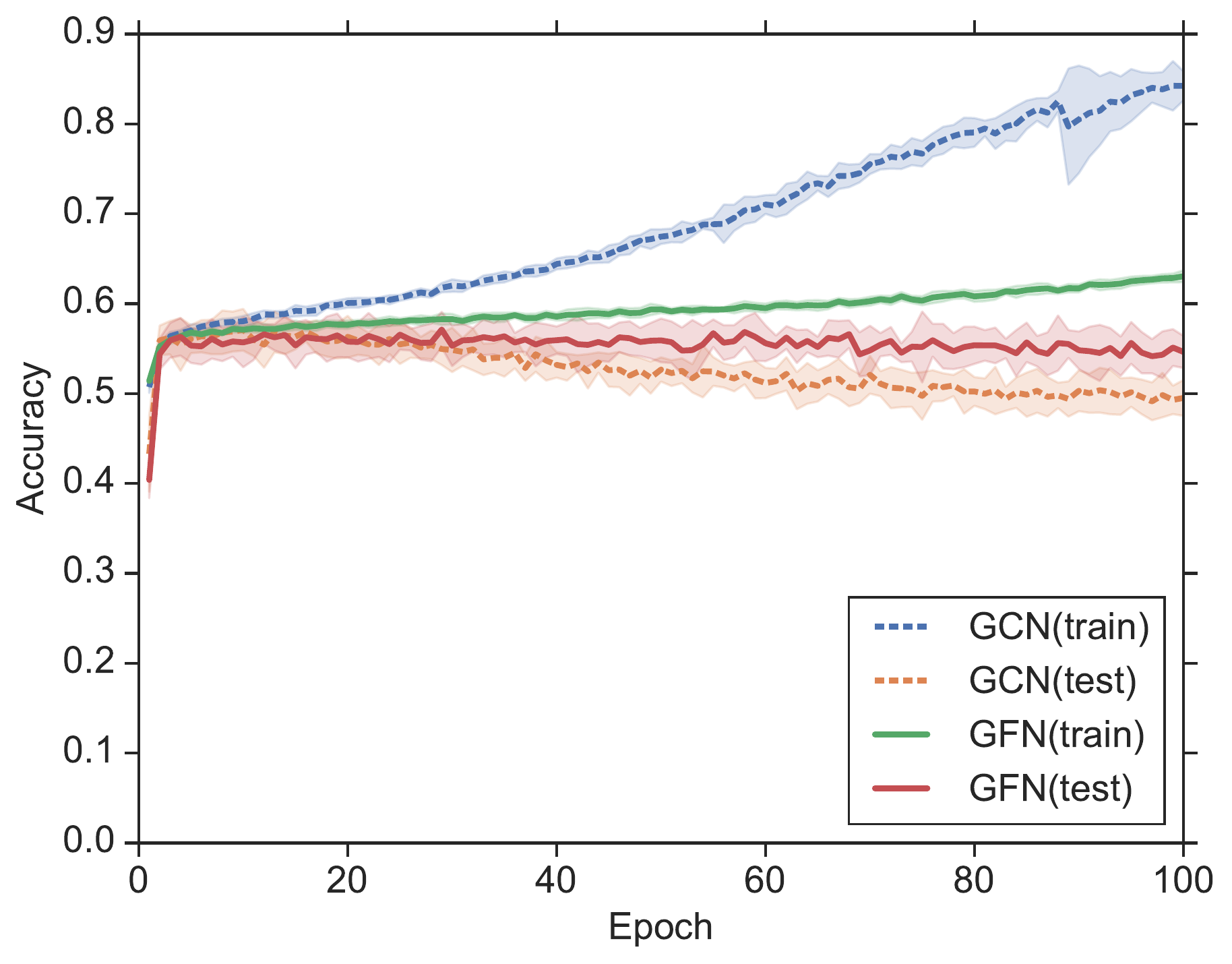}
        \caption{REDDIT-MULTI-5K}
    \end{subfigure}
\caption{\label{fig:perf_vs_epoch_more} Training and test performance versus training epoch.}
\end{figure}

Figure \ref{fig:perf_vs_epoch_more} shows more training/test curves for both GCN and GFN. The conclusion is consistent with main text that GFN works well not because it is easier to optimize.

\section{Detailed performances with different features}

\begin{table}[!htbp]
\small
\centering
\caption{Accuracies (\%) under various augmented features. $A^{1..3}X$ is abbreviated for $A^1X,A^2X,A^3X$, and default node feature $X$ is always used (if available) but not displayed to reduce clutter.}
\label{tab:features_more}
\begin{tabular}{ll|c|cccc|ccc}
\Xhline{1.6\arrayrulewidth}\toprule
Dataset       & Model &   None &  $\bm d$  &  $A^1X$ &   $A^{1,2}X$ &   $A^{1..3}X$ &  $\bm d, A^1X$ &   $\bm d, A^{1,2}X$ & $\bm d,A^{1..3}X$ \\
\midrule\midrule
\multirow{2}{*}{MUTAG} & GCN &  83.48 &     87.09 &  83.35 &  83.43 &  85.56 &         87.18 &         87.62 &         88.73 \\
                 & GFN &  82.21 &     89.31 &  87.59 &  87.17 &  86.62 &         89.42 &         89.28 &         88.26 \\
\midrule
\multirow{2}{*}{NCI1} & GCN &  80.15 &     83.24 &  82.62 &  83.11 &  82.60 &         83.38 &         83.63 &         83.50 \\
                 & GFN &  70.83 &     75.50 &  80.95 &  82.80 &  83.50 &         81.92 &         82.41 &         82.84 \\
\midrule
\multirow{2}{*}{PROTEINS} & GCN &  74.49 &     76.28 &  74.48 &  75.47 &  76.54 &         77.09 &         76.91 &         77.45 \\
                 & GFN &  74.93 &     76.63 &  76.01 &  75.74 &  76.64 &         76.37 &         76.46 &         77.09 \\
\midrule
\multirow{2}{*}{DD} & GCN &  79.29 &     78.78 &  78.70 &  77.67 &  78.18 &         78.35 &         78.79 &         79.12 \\
                 & GFN &  78.70 &     77.77 &  77.85 &  77.43 &  78.28 &         77.34 &         76.92 &         78.11 \\
\midrule
\multirow{2}{*}{ENZYMES} & GCN &  75.17 &     67.17 &  72.00 &  71.50 &  70.50 &         69.50 &         69.33 &         69.67 \\
                 & GFN &  74.67 &     70.00 &  71.50 &  72.33 &  70.83 &         68.50 &         71.00 &         69.33 \\
\midrule
\multirow{2}{*}{COLLAB} & GCN &  39.69 &     82.14 &  76.62 &  76.98 &  77.22 &         82.14 &         82.24 &         82.20 \\
                 & GFN &  31.57 &     80.36 &  76.40 &  77.08 &  77.04 &         81.28 &         81.62 &         81.26 \\
\midrule
\multirow{2}{*}{IMDB-B} & GCN &  51.00 &     73.00 &  70.30 &  71.10 &  72.20 &         73.50 &         73.80 &         73.70 \\
                 & GFN &  50.00 &     73.30 &  72.30 &  71.30 &  71.70 &         74.40 &         73.20 &         73.90 \\
\midrule
\multirow{2}{*}{IMDB-M} & GCN &  35.00 &     50.33 &  45.53 &  46.33 &  45.73 &         50.20 &         50.73 &         51.00 \\
                 & GFN &  33.33 &     51.20 &  46.80 &  46.67 &  46.47 &         51.93 &         51.93 &         51.73 \\
\midrule
\multirow{2}{*}{RE-M5K} & GCN &  28.48 &     56.99 &  54.97 &  57.43 &  56.55 &         56.67 &         56.75 &         57.01 \\
                 & GFN &  20.00 &     54.23 &  51.11 &  55.85 &  56.35 &         56.45 &         57.01 &         56.71 \\
\midrule
\multirow{2}{*}{RE-M12K} & GCN &  15.93 &     49.28 &  48.58 &  50.11 &  49.71 &         49.73 &         50.03 &         49.92 \\
                 & GFN &  17.33 &     44.86 &  43.61 &  48.25 &  48.87 &         48.31 &         49.37 &         49.39 \\
\bottomrule\Xhline{1.6\arrayrulewidth}
\end{tabular}
\end{table}

Table \ref{tab:features_more} show the performances under different graph features for GNNs and GFNs. It is evident that both model benefit significantly from graph features, especially GFNs.

\section{Detailed performances with different architecture depths}

Table \ref{tab:layers_more} shows performance per datasets under different number of layers.

\begin{table}[!htbp]
\small
\centering
\caption{Accuracies (\%) under different number of Conv. layers. Flat denotes the collapsed GFN into a linear model (i.e. linearizing the set function).}
\label{tab:layers_more}
\begin{tabular}{ll|c|ccccc}
\Xhline{1.6\arrayrulewidth}\toprule
Dataset & Method & Flat &     1 &      2 &      3 &      4 &      5\\
\midrule\midrule
\multirow{2}{*}{MUTAG} & GCN &    - &  88.32 &  90.89 &  87.65 &  88.31 &  87.68 \\
                 & GFN &  82.85 &  90.34 &  89.39 &  88.18 &  87.59 &  87.18 \\
\midrule
\multirow{2}{*}{NCI1} & GCN &    - &  75.62 &  81.41 &  83.04 &  82.94 &  83.31 \\
                 & GFN &  68.61 &  81.77 &  83.09 &  82.85 &  82.80 &  83.09 \\
\midrule
\multirow{2}{*}{PROTEINS} & GCN &    - &  76.91 &  76.99 &  77.00 &  76.19 &  75.29 \\
                 & GFN &  75.65 &  77.71 &  77.09 &  77.17 &  76.28 &  75.92 \\
\midrule
\multirow{2}{*}{DD} & GCN &    - &  77.34 &  77.93 &  78.95 &  79.46 &  78.77 \\
                 & GFN &  76.75 &  78.44 &  78.78 &  79.04 &  78.45 &  76.32 \\
\midrule
\multirow{2}{*}{ENZYMES} & GCN &    - &  67.67 &  69.67 &  67.67 &  66.83 &  66.00 \\
                 & GFN &  43.83 &  69.67 &  70.50 &  71.67 &  69.83 &  68.17 \\
\midrule
\multirow{2}{*}{COLLAB} & GCN &    - &  80.36 &  81.86 &  81.40 &  81.90 &  81.78 \\
                 & GFN &  75.72 &  81.24 &  82.04 &  81.36 &  82.18 &  81.72 \\
\midrule
\multirow{2}{*}{IMDB-B} & GCN &    - &  72.60 &  72.30 &  73.30 &  73.80 &  73.40 \\
                 & GFN &  73.10 &  73.50 &  73.30 &  74.00 &  73.90 &  73.60 \\
\midrule
\multirow{2}{*}{IMDB-M} & GCN &    - &  51.53 &  51.07 &  50.87 &  51.53 &  50.60 \\
                 & GFN &  50.40 &  51.73 &  52.13 &  51.93 &  51.87 &  51.40 \\
\midrule
\multirow{2}{*}{RE-M5K} & GCN &    - &  54.05 &  56.49 &  56.83 &  56.73 &  56.89 \\
                 & GFN &  52.97 &  57.45 &  57.13 &  57.21 &  56.61 &  57.03 \\
\midrule
\multirow{2}{*}{RE-M12K} & GCN &    - &  44.91 &  48.87 &  49.45 &  49.52 &  49.61 \\
                 & GFN &  39.84 &  49.58 &  49.82 &  49.54 &  49.44 &  49.27 \\
\bottomrule\Xhline{1.6\arrayrulewidth}
\end{tabular}
\end{table}

\section{Graph visualizations}
\label{sec:appx_vis}
\begin{figure*}[!h]
    \centering
    \begin{subfigure}[b]{0.48\textwidth}
        \includegraphics[width=\textwidth]{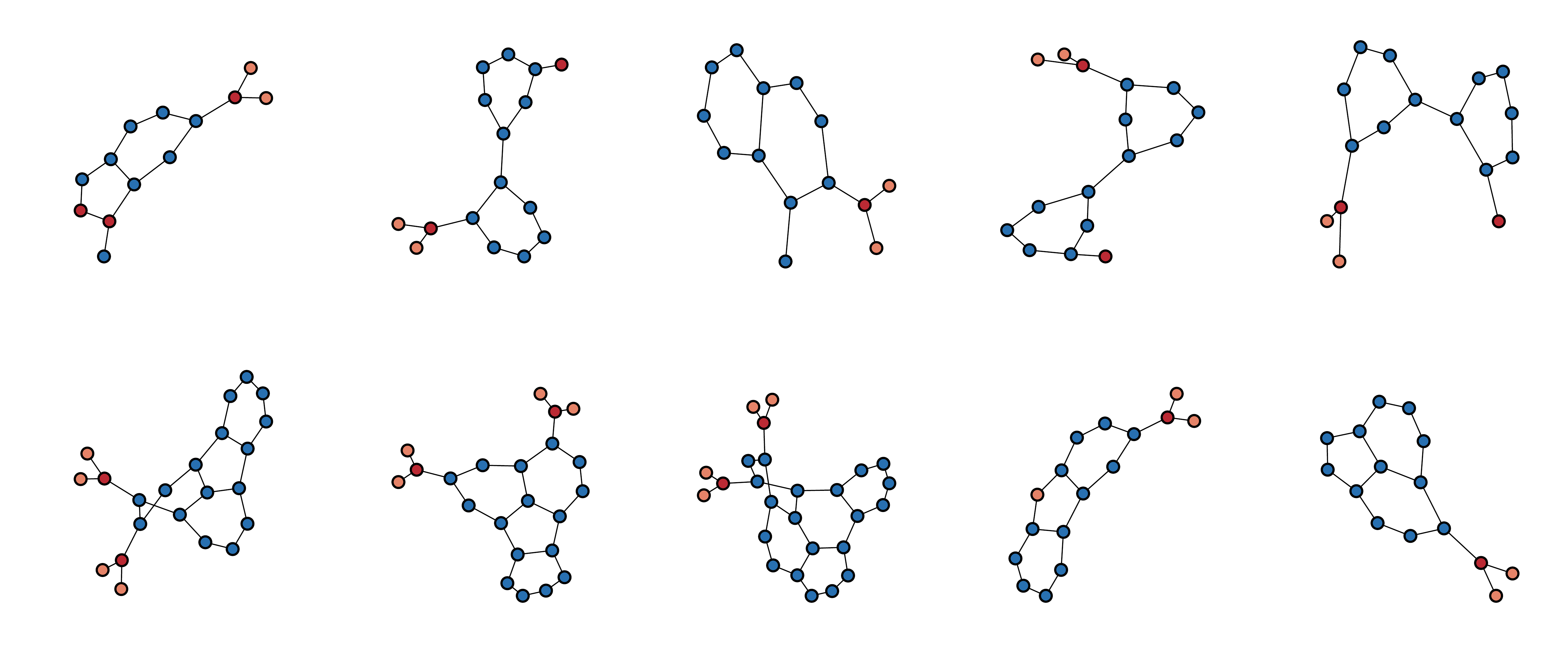}
        \caption{Random samples.}
    \end{subfigure}
    ~~~
    \begin{subfigure}[b]{0.48\textwidth}
        \includegraphics[width=\textwidth]{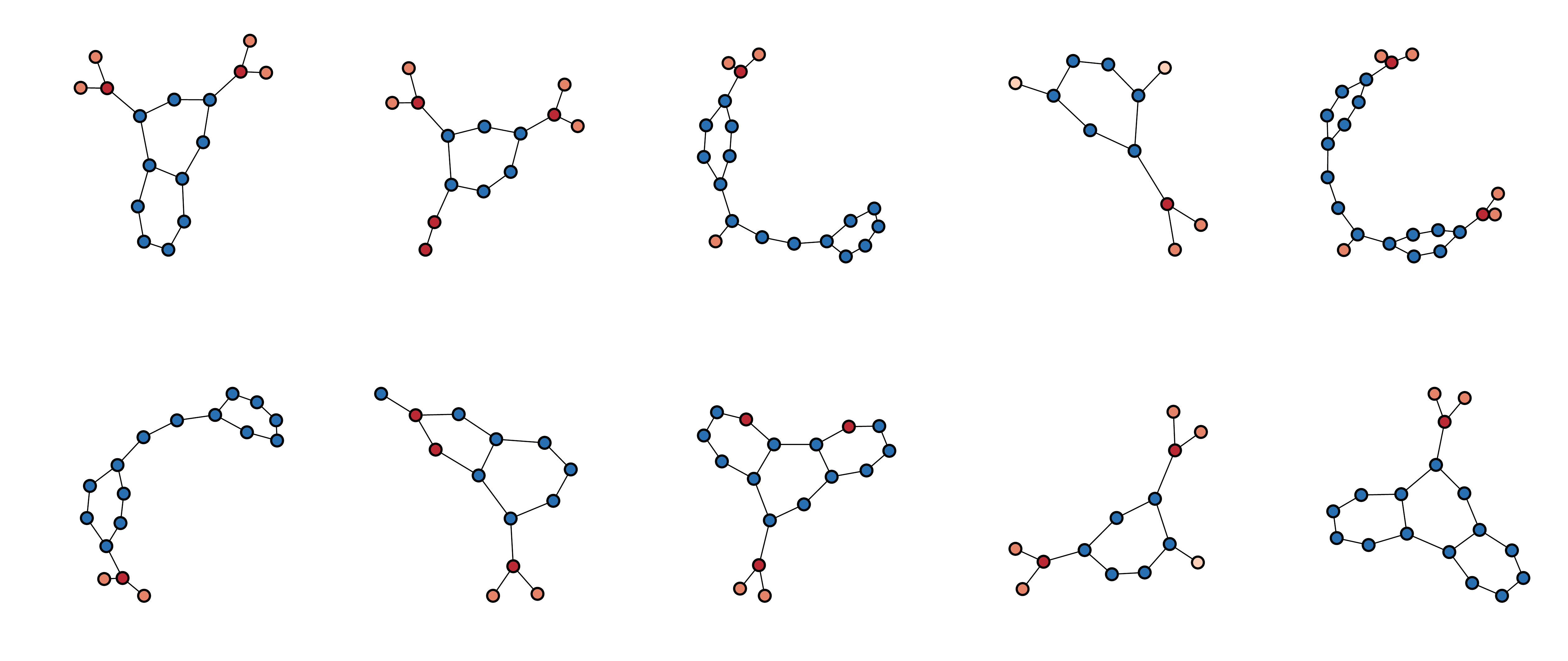}
        \caption{Mis-classified samples by GFN.}
    \end{subfigure}
    \begin{subfigure}[b]{0.48\textwidth}
        \includegraphics[width=\textwidth]{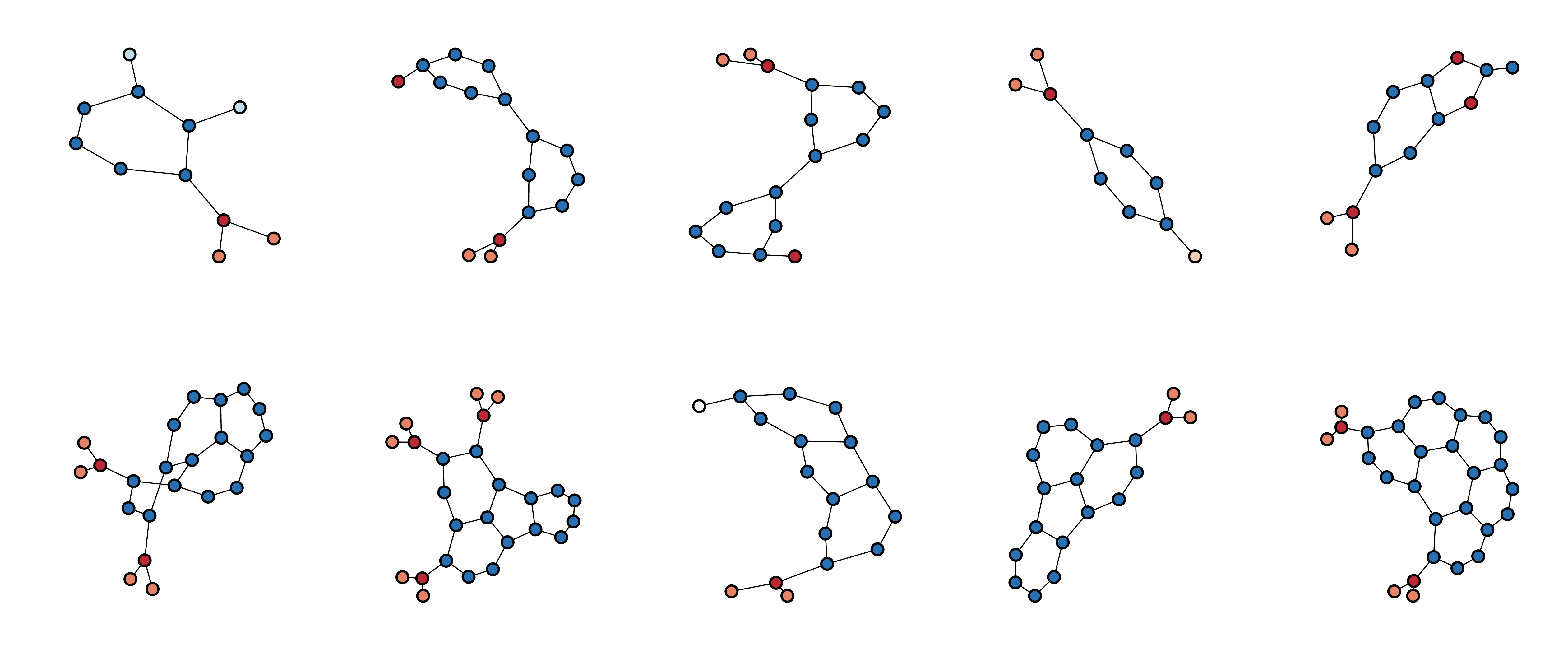}
        \caption{Random samples.}
    \end{subfigure}
    ~~~
    \begin{subfigure}[b]{0.48\textwidth}
        \includegraphics[width=\textwidth]{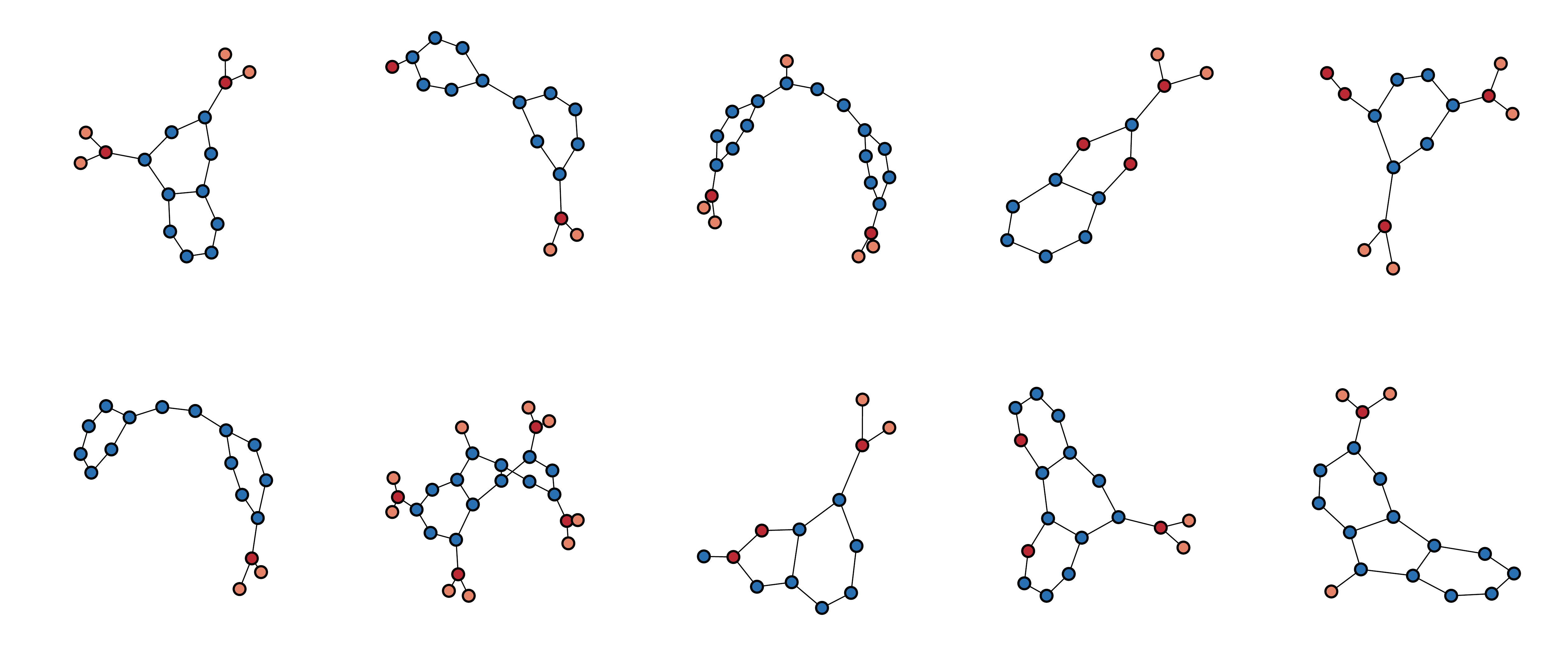}
        \caption{Mis-classified samples by GCN.}
    \end{subfigure}
    \caption{Random and mis-classified samples from MUTAG. Each row represents a (true) class.}\label{fig:random_mis_mutag}
\end{figure*}

\begin{figure*}[!h]
    \centering
    \begin{subfigure}[b]{0.48\textwidth}
        \includegraphics[width=\textwidth]{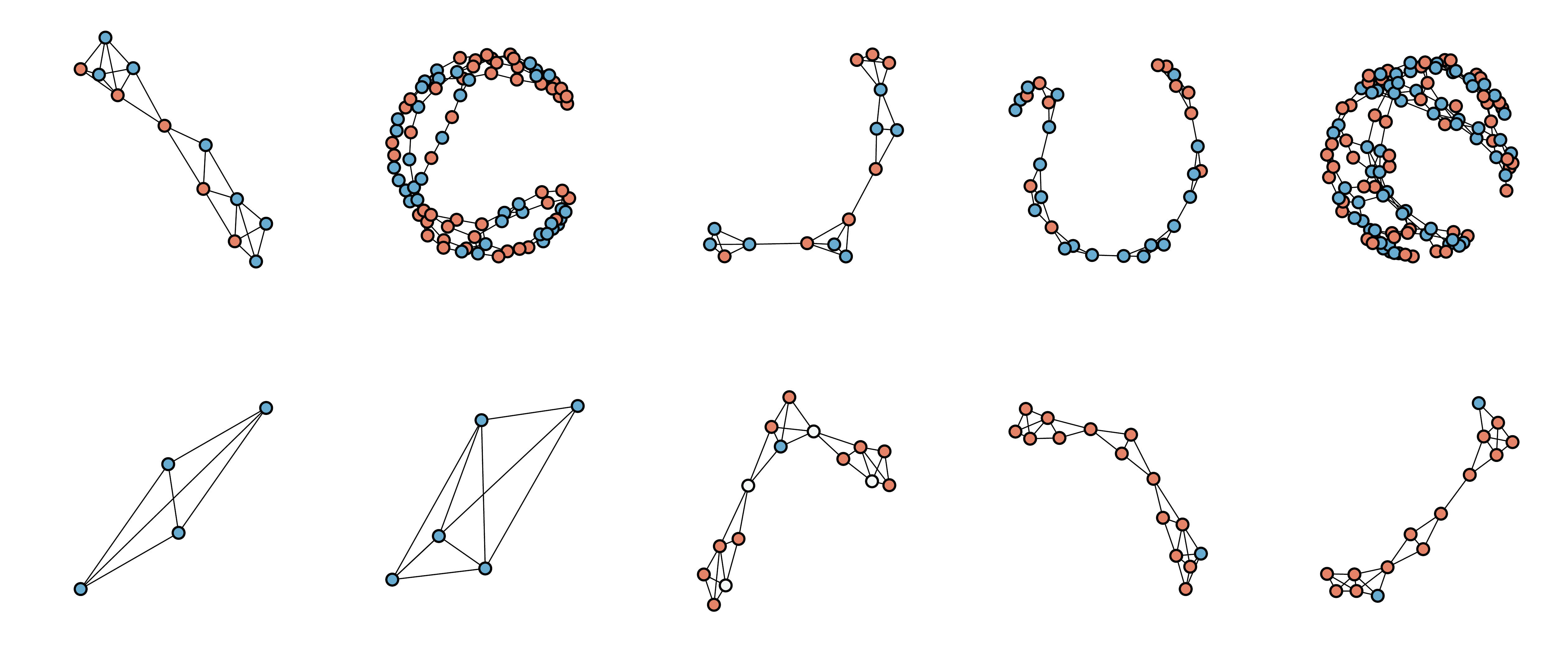}
        \caption{Random samples}
    \end{subfigure}
    ~~~
    \begin{subfigure}[b]{0.48\textwidth}
        \includegraphics[width=\textwidth]{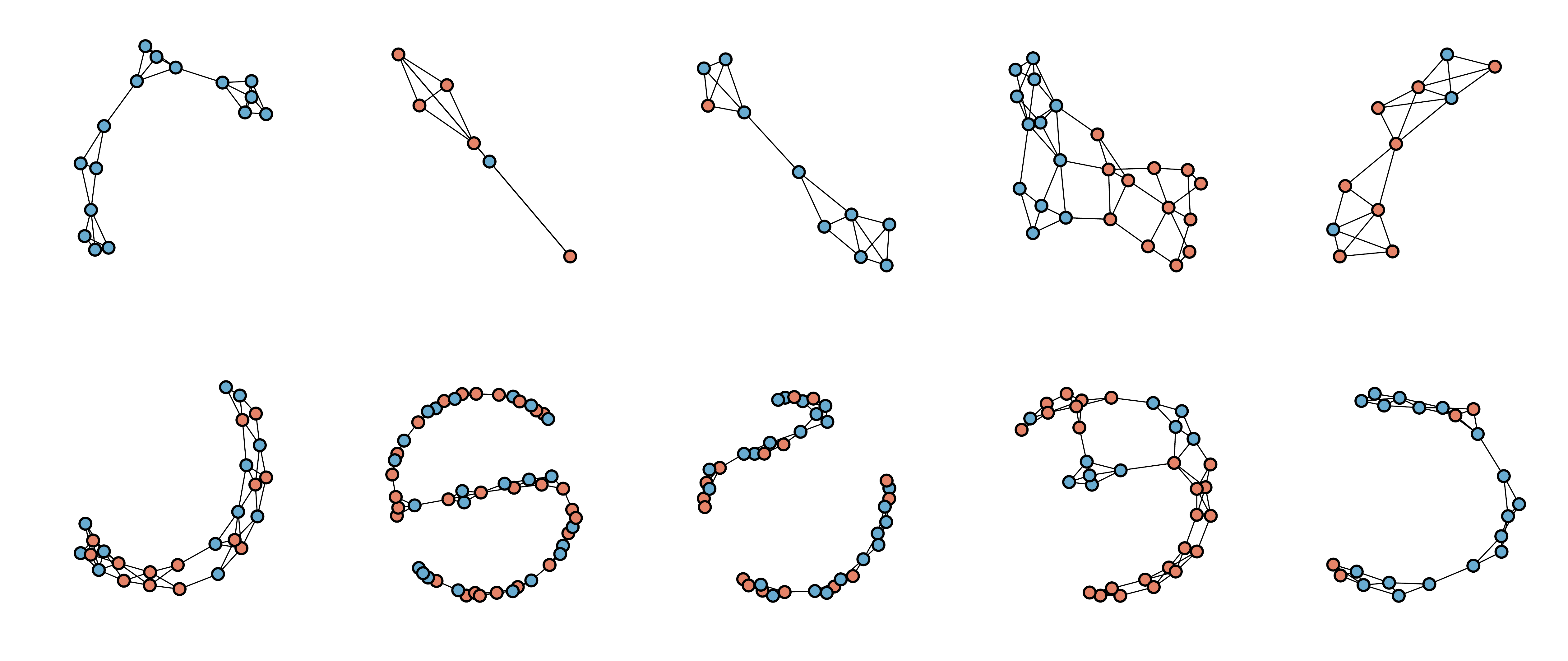}
        \caption{Mis-classified samples by GFN.}
    \end{subfigure}
    \begin{subfigure}[b]{0.48\textwidth}
        \includegraphics[width=\textwidth]{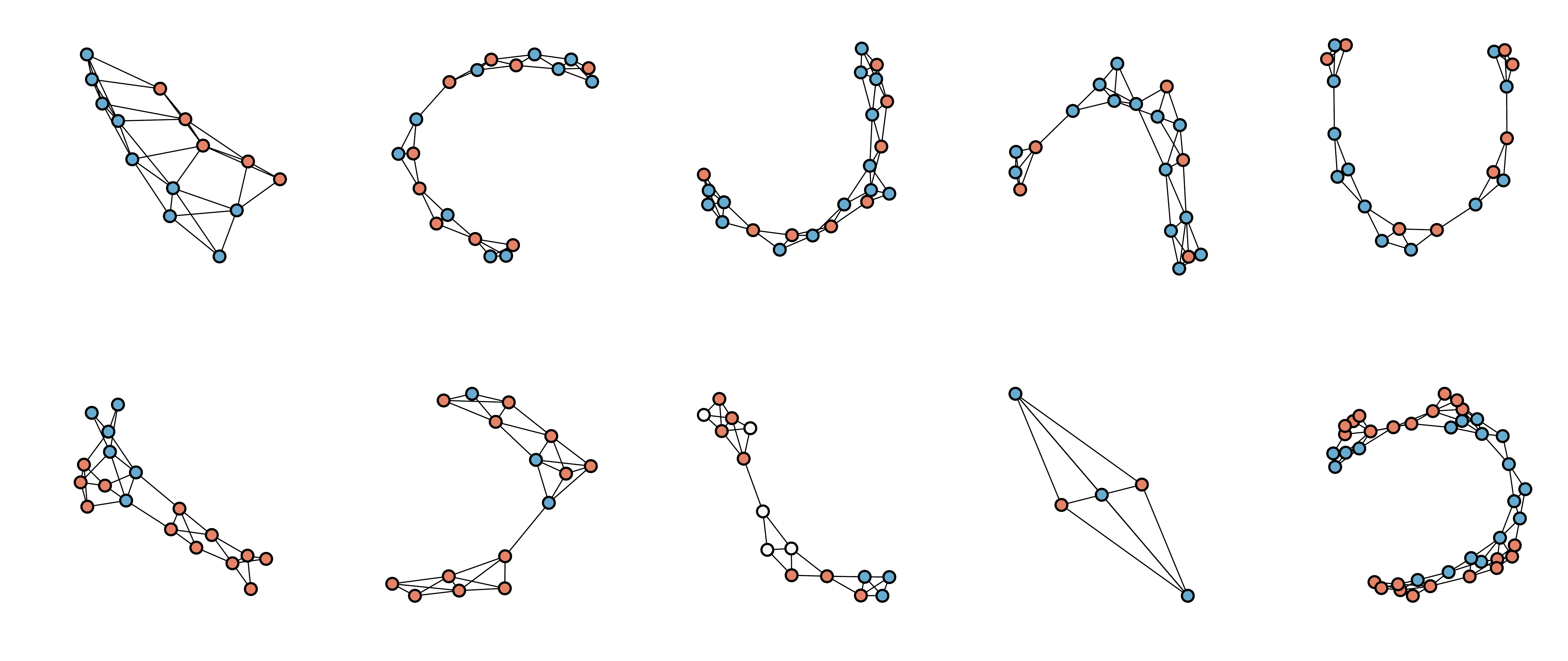}
        \caption{Random samples.}
    \end{subfigure}
    ~~~
    \begin{subfigure}[b]{0.48\textwidth}
        \includegraphics[width=\textwidth]{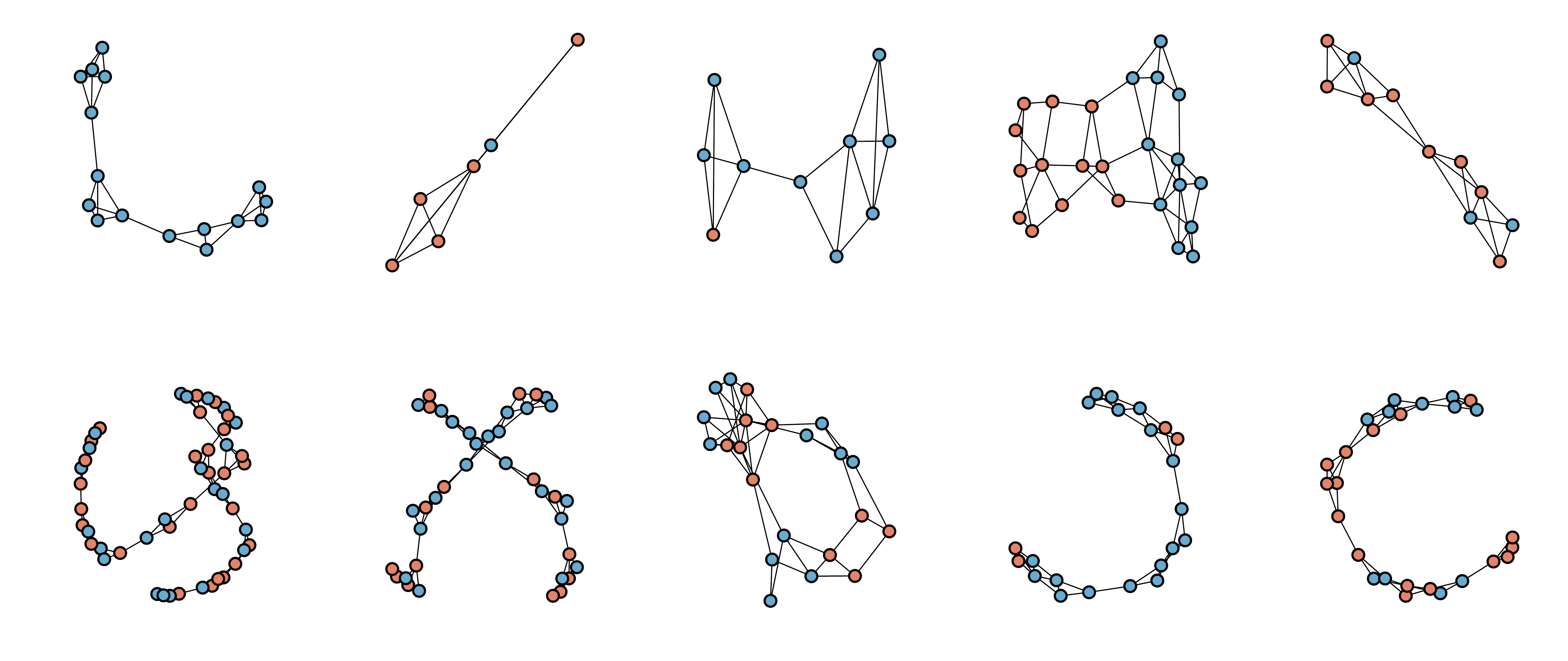}
        \caption{Mis-classified samples by GCN.}
    \end{subfigure}
    \caption{Random and mis-classified samples from PROTEINS. Each row represents a (true) class.}\label{fig:random_mis_proteins}
\end{figure*}

\begin{figure*}[!t]
    \centering
    \begin{subfigure}[b]{0.48\textwidth}
        \includegraphics[width=\textwidth]{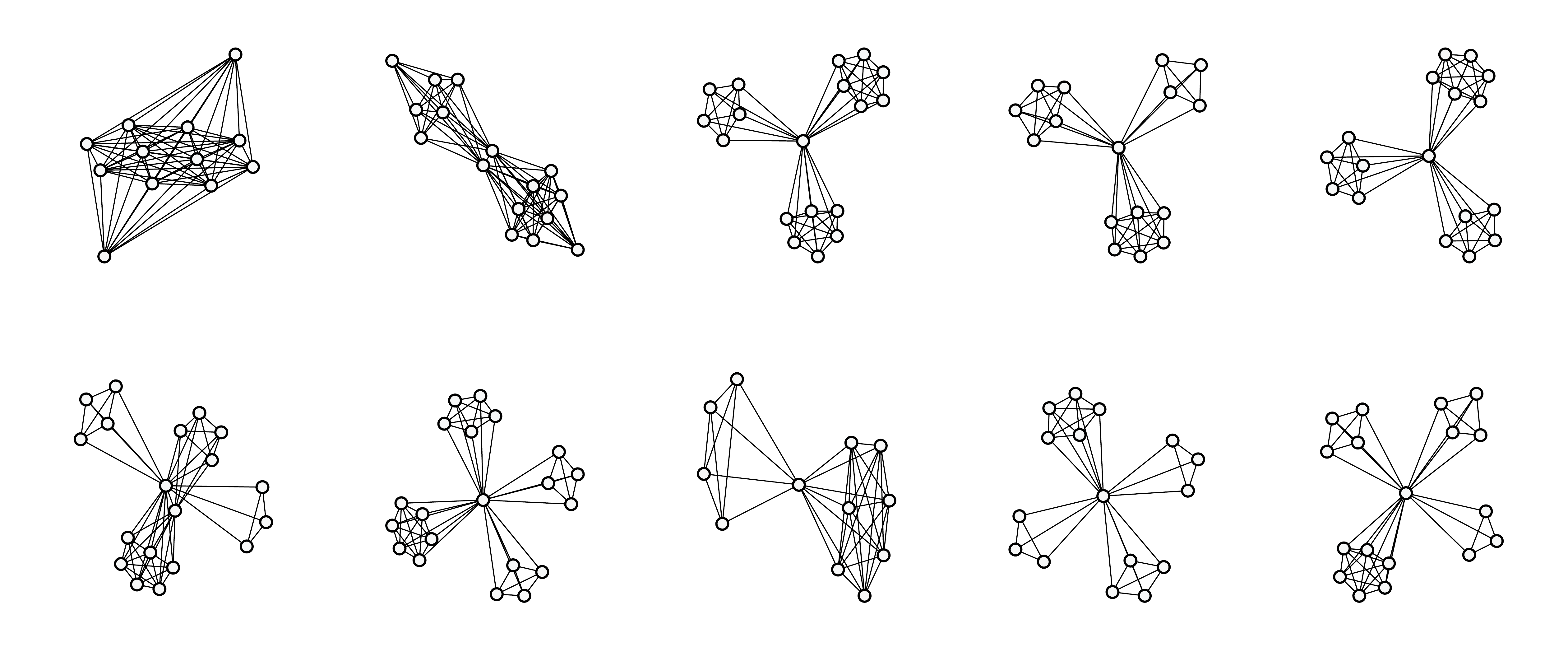}
        \caption{Random samples.}
    \end{subfigure}
    ~~~
    \begin{subfigure}[b]{0.48\textwidth}
        \includegraphics[width=\textwidth]{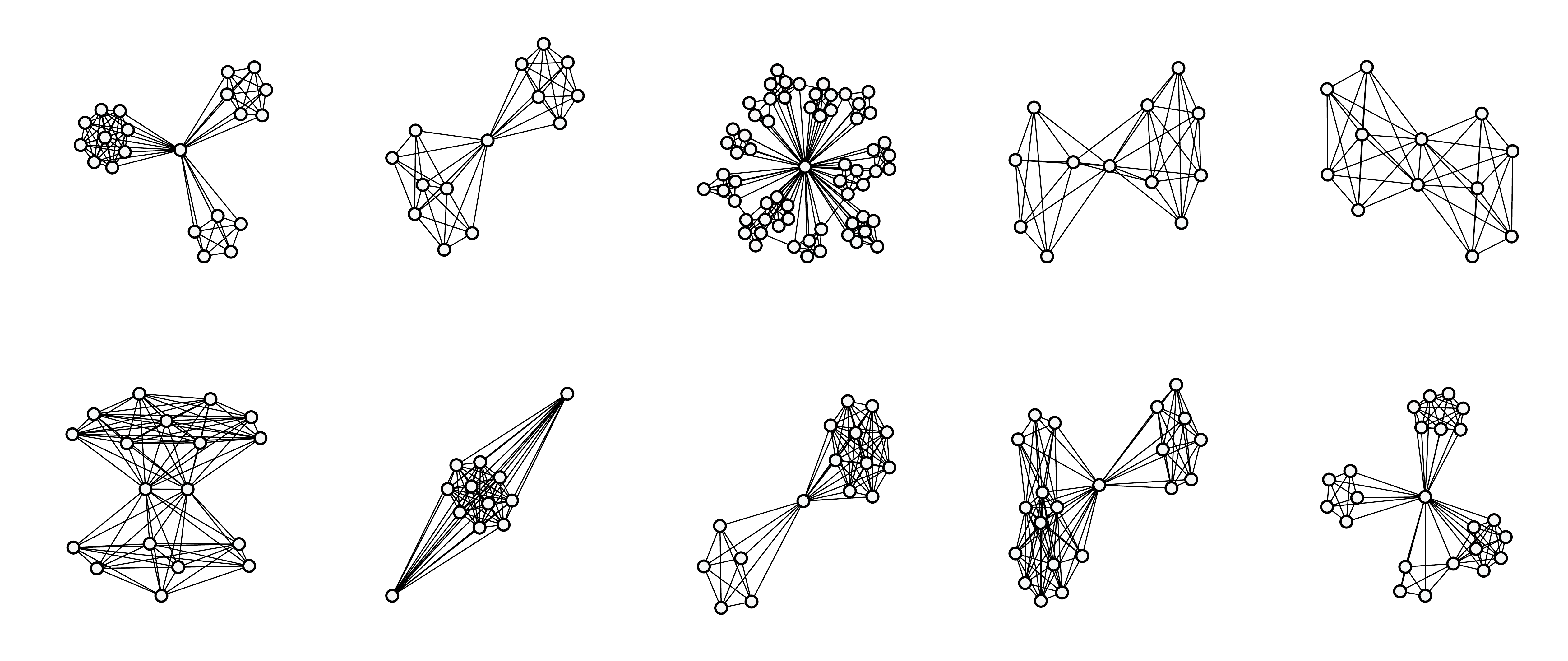}
        \caption{Mis-classified samples by GFN.}
    \end{subfigure}
    \begin{subfigure}[b]{0.48\textwidth}
        \includegraphics[width=\textwidth]{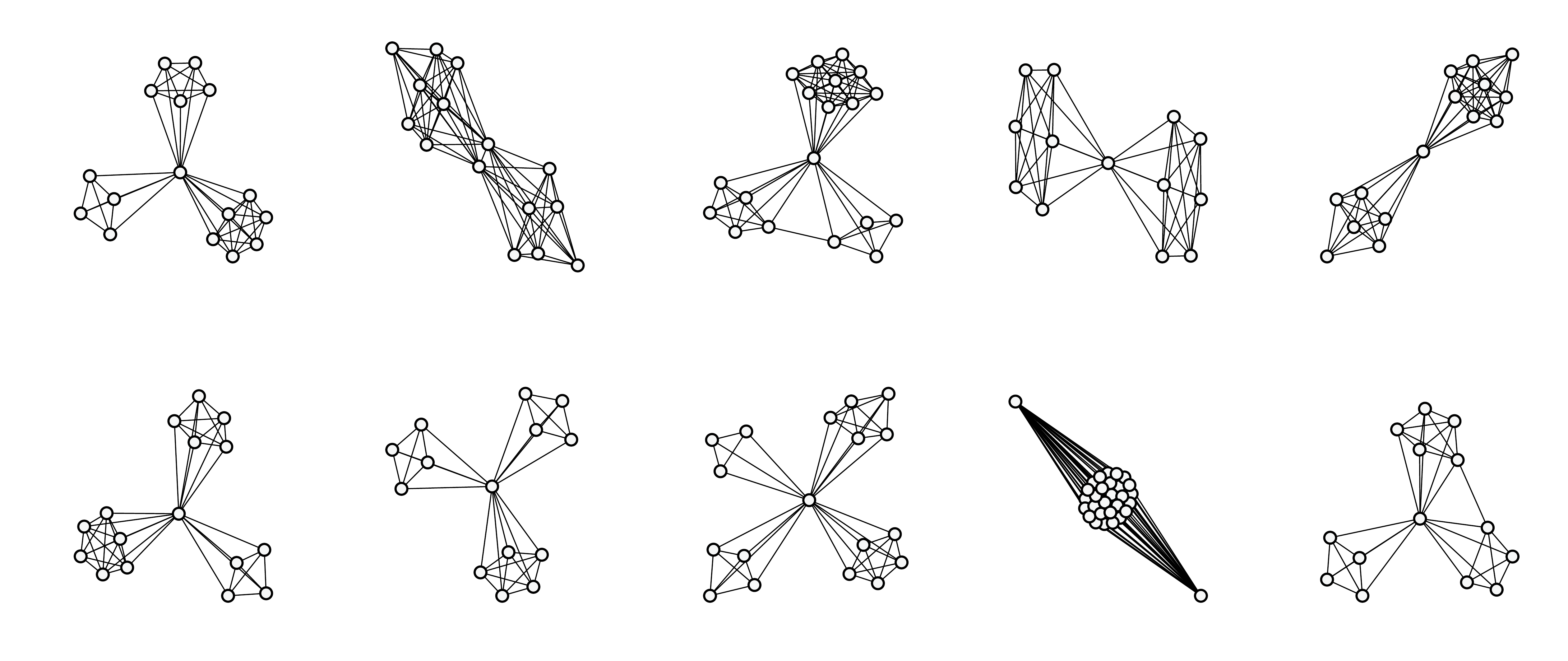}
        \caption{Random samples.}
    \end{subfigure}
    ~~~
    \begin{subfigure}[b]{0.48\textwidth}
        \includegraphics[width=\textwidth]{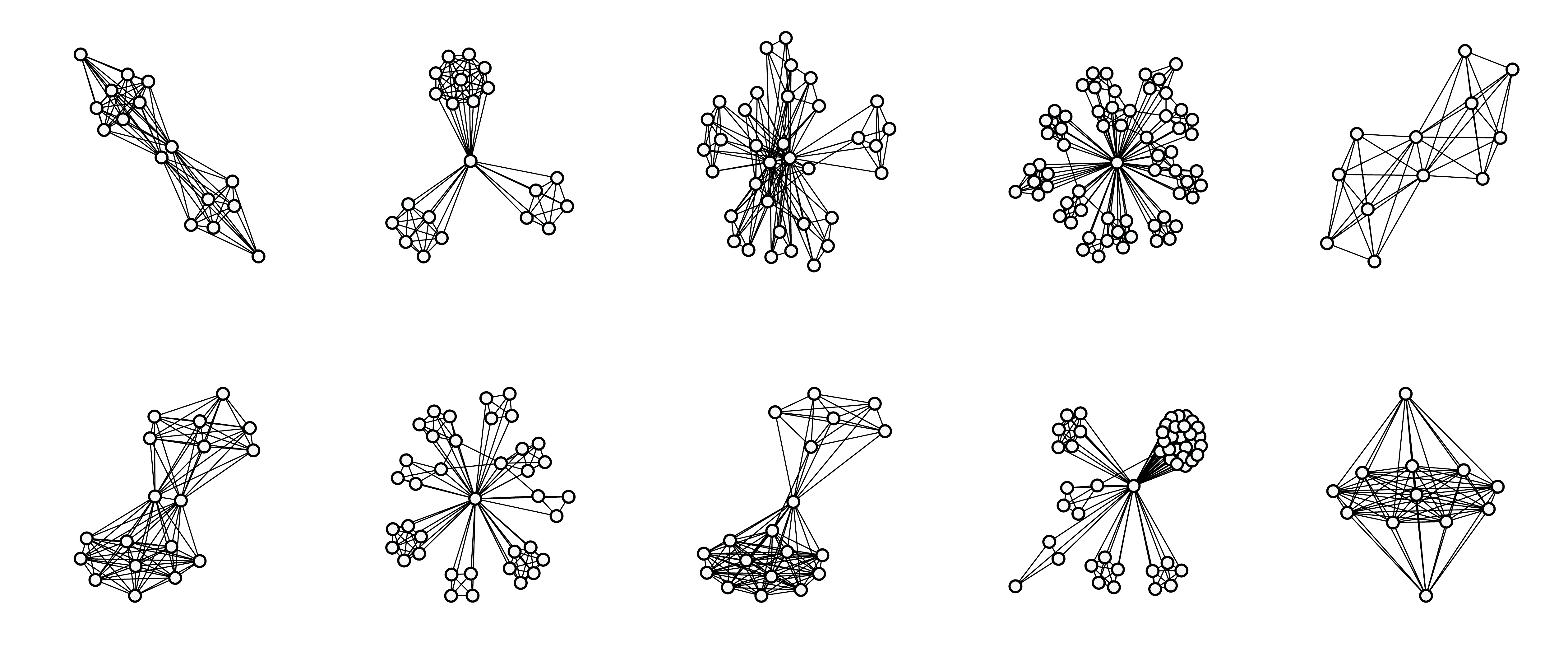}
        \caption{Mis-classified samples by GCN.}
    \end{subfigure}
    \caption{Random and mis-classified samples from IMDB-B. Each row represents a (true) class.}\label{fig:random_mis_imdb-m}
\end{figure*}

\begin{figure*}[!t]
    \centering
    \begin{subfigure}[b]{0.48\textwidth}
        \includegraphics[width=\textwidth]{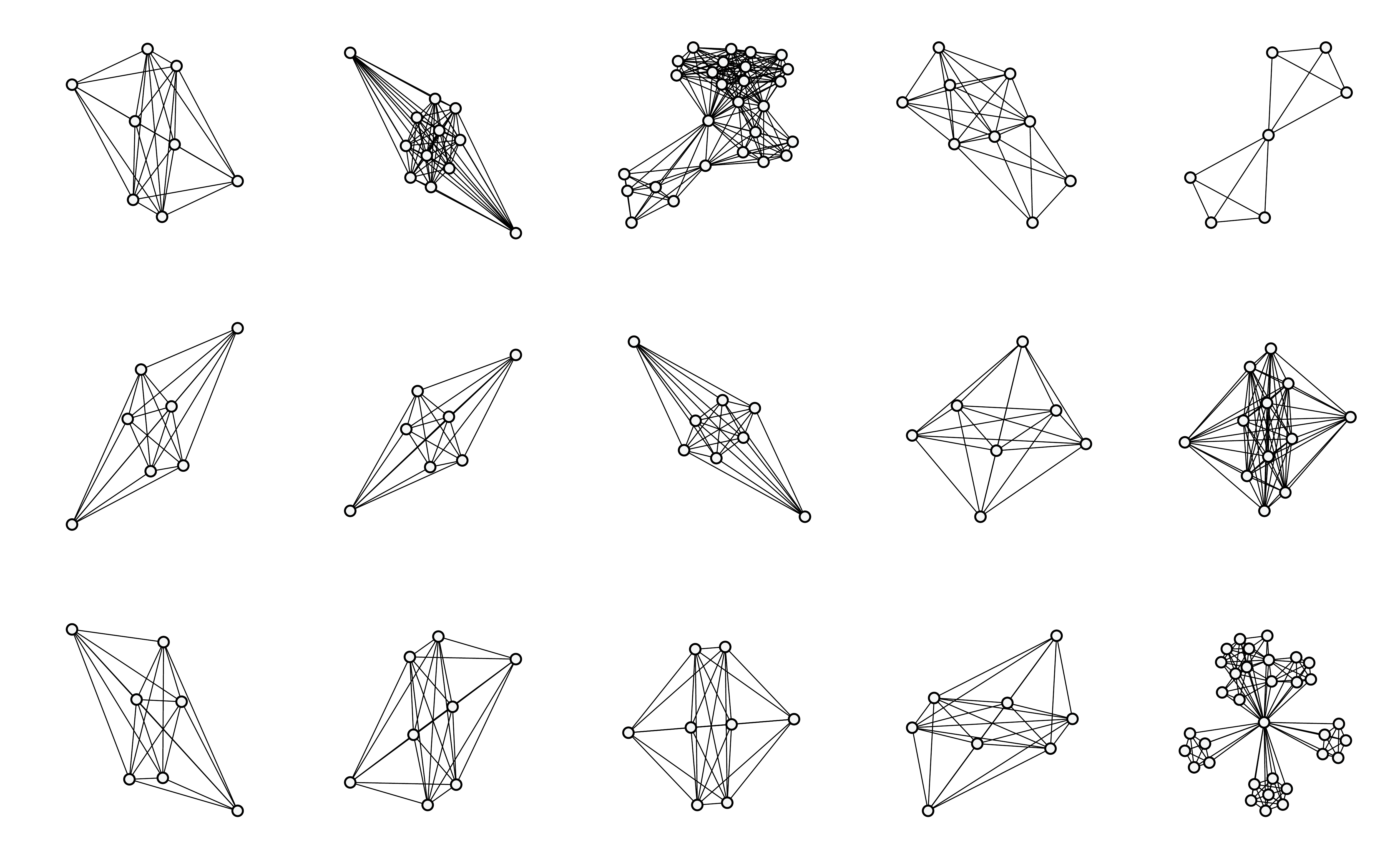}
        \caption{Random samples.}
    \end{subfigure}
    ~~~
    \begin{subfigure}[b]{0.48\textwidth}
        \includegraphics[width=\textwidth]{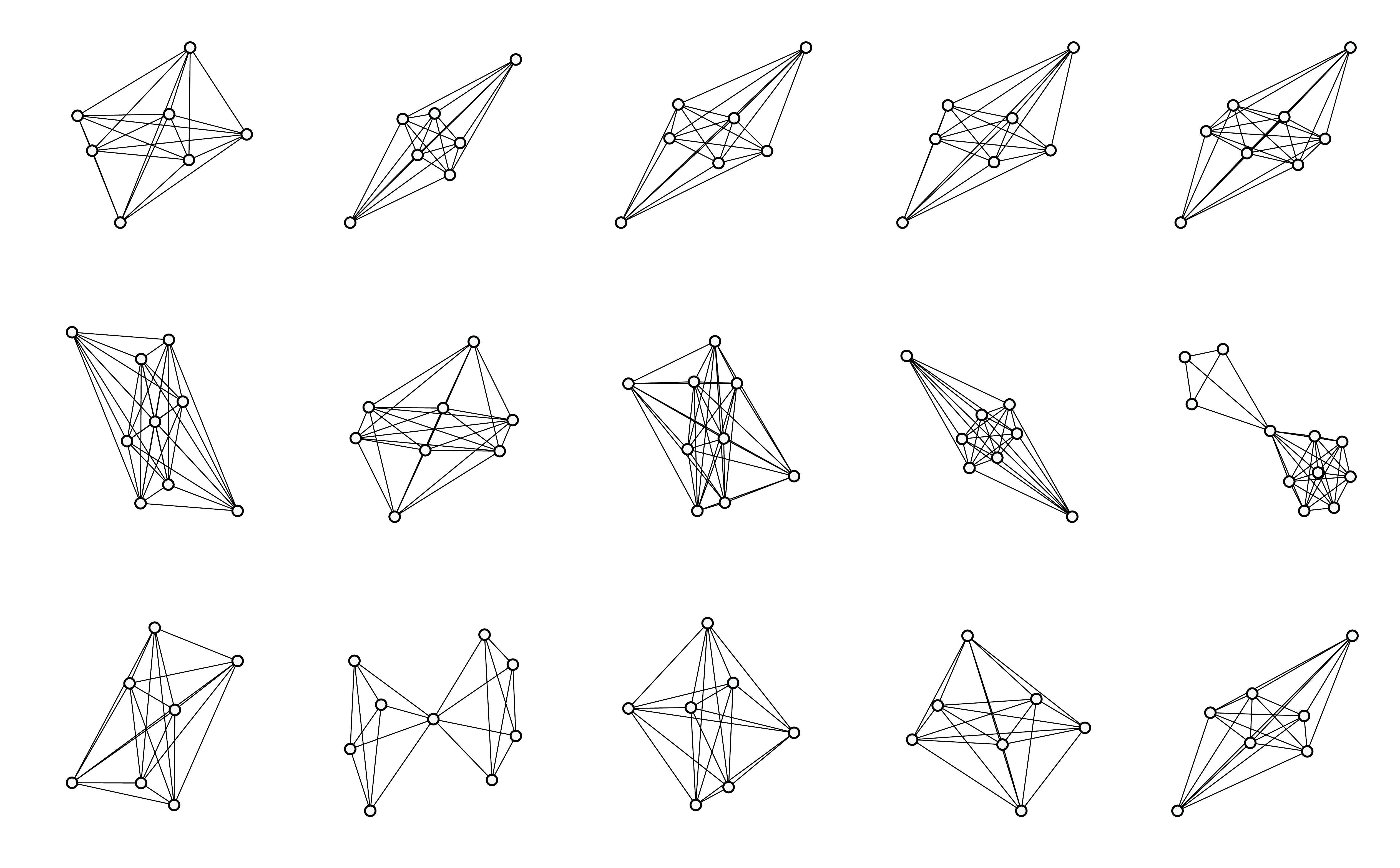}
        \caption{Mis-classified samples by GFN.}
    \end{subfigure}
    \begin{subfigure}[b]{0.48\textwidth}
        \includegraphics[width=\textwidth]{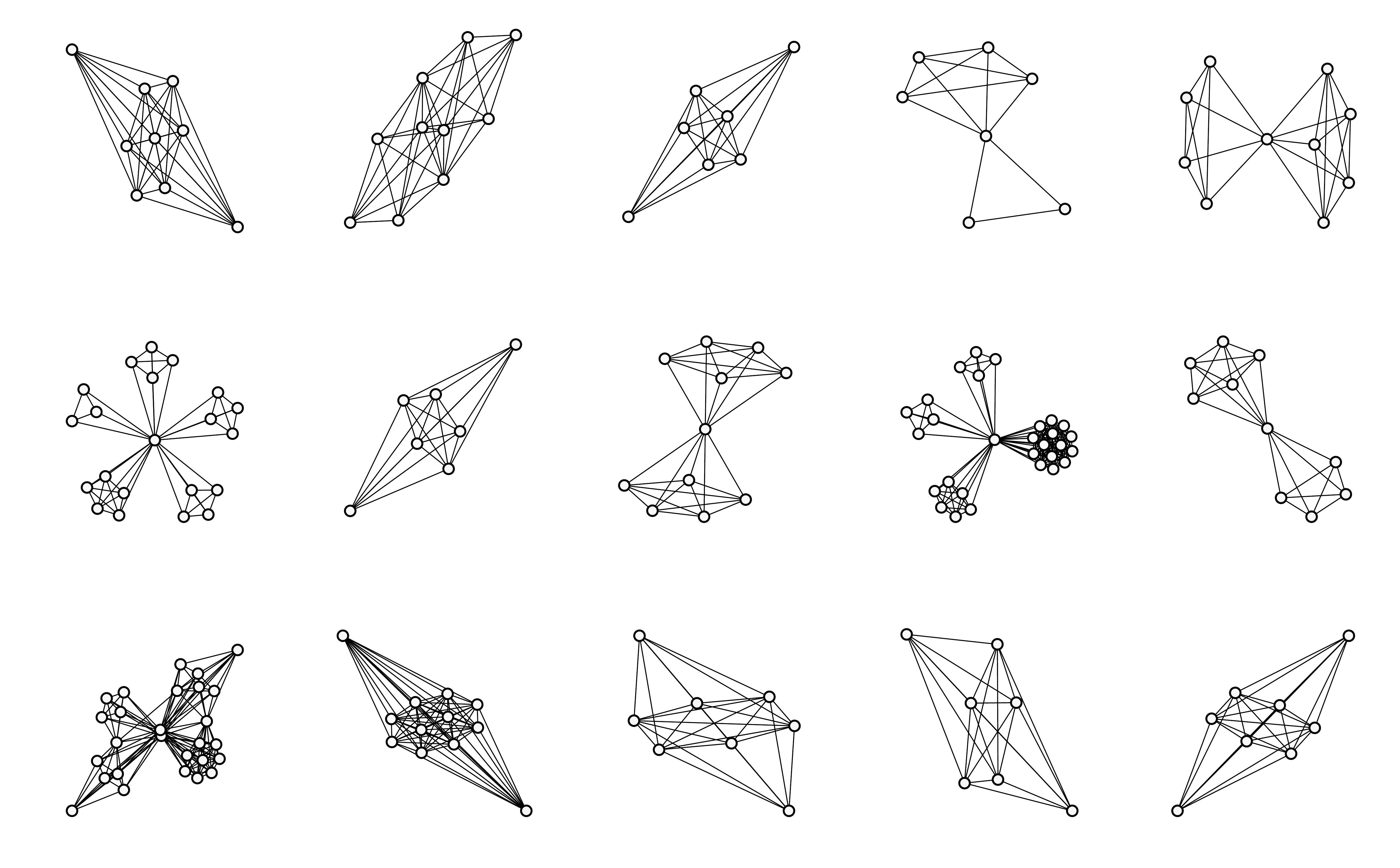}
        \caption{Random samples.}
    \end{subfigure}
    ~~~
    \begin{subfigure}[b]{0.48\textwidth}
        \includegraphics[width=\textwidth]{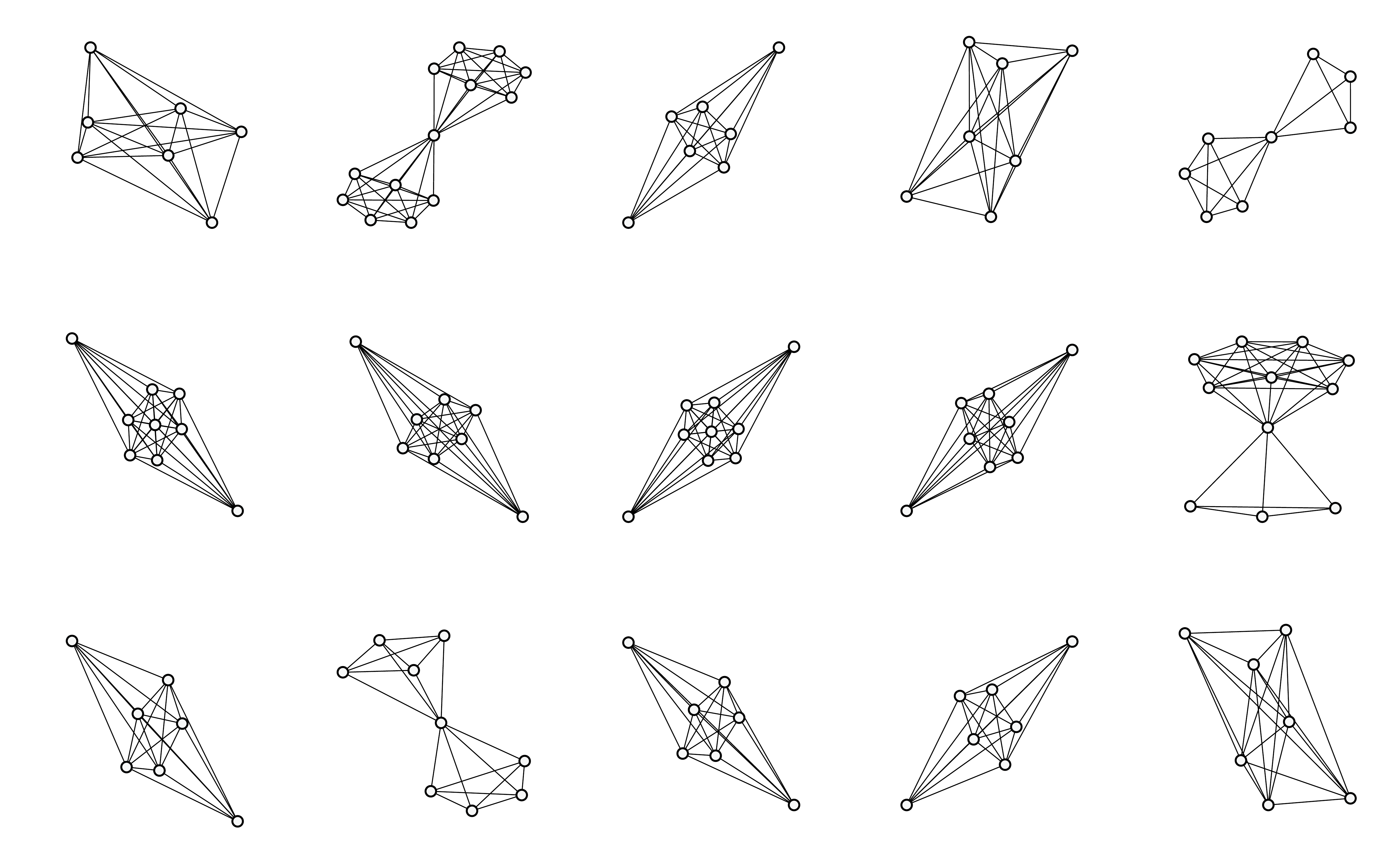}
        \caption{Mis-classified samples by GCN.}
    \end{subfigure}
    \caption{Random and mis-classified samples from IMDB-M. Each row represents a (true) class.}\label{fig:random_mis_imdb-b}
\end{figure*}

Figure \ref{fig:random_mis_mutag}, \ref{fig:random_mis_proteins}, \ref{fig:random_mis_imdb-b}, and \ref{fig:random_mis_imdb-m} show the random and mis-classified samples for MUTAG, PROTEINS, IMDB-B, and IMDB-M, respectively. In general, it is difficult to find the patterns of each class by visually examining the graphs. And the mis-classified patterns are not visually distinguishable, except for IMDB-B/IMDB-M datasets where there are some graphs seem ambiguous.

\end{document}